\documentclass{article}

     \usepackage[preprint, nonatbib]{neurips_2020}

\usepackage[utf8]{inputenc} 
\usepackage[T1]{fontenc}    
\usepackage{hyperref}       
\usepackage{url}            
\usepackage{booktabs}       
\usepackage{amsfonts}       
\usepackage{nicefrac}       
\usepackage{microtype}      
\usepackage{amsmath}
\usepackage{accents}
\usepackage{afterpage}
\usepackage{titlesec}
\usepackage{graphicx} 
\usepackage{tabu}                      
\usepackage[noend]{algpseudocode}
\usepackage{algorithm,algorithmicx}
\usepackage{subcaption}
\usepackage{textcomp}

\newcommand*\Let[2]{\State #1 $\gets$ #2}
\algrenewcommand\algorithmicrequire{\textbf{Precondition:}}
\algrenewcommand\algorithmicensure{\textbf{Postcondition:}}

\title{Extending DeepSDF for automatic 3D shape retrieval and similarity transform estimation.}

\author{%
    Oladapo A. Afolabi,
    Allen Y. Yang,
    S. Shankar Sastry\\
    Department of Electrical Engineering and Computer Sciences\\
    University of California, Berkeley\\
    \texttt{\{oafolabi,yang,sastry\}@eecs.berkeley.edu} \\
}

\begin{document}

\maketitle

\begin{abstract}
    \vspace{-0.2cm}
Recent advances in computer graphics and computer vision have found successful application of deep neural network models for 3D shapes based on signed distance functions (SDFs) that are useful for shape representation, retrieval, and completion. However, this approach has been limited by the need to have query shapes in the same canonical scale and pose as those observed during training, restricting its effectiveness on real world scenes. We present a formulation to overcome this issue by jointly estimating shape and similarity transform parameters. We conduct experiments to demonstrate the effectiveness of this formulation on synthetic and real datasets and report favorable comparisons to the state of the art. Finally, we also emphasize the viability of this approach as a form of 3D model compression.
\end{abstract}

\vspace{-0.3cm}
\section{Introduction}
\vspace{-0.3cm}
The rapid growth of low-cost LIDAR or RGB-D sensors (e.g., Intel RealSense and Apple iPad Pro LIDAR) has led to the growth of applications that require building 3D models of real-world scenes. Such models are useful for virtual immersion and automatic content creation in Augmented and Virtual Reality (AR/VR) where the user experience can be customized to individuals' own spaces. A typical pipeline of the task currently would include Simultaneous Localization and Mapping to generate an SDF representation of the space. Subsequently, mesh or pointcloud models of the scene may be extracted from the SDFs using variants of the Marching Cubes algorithm\cite{lorensen1987marching}. 

While very accurate models can be produced by this pipeline, it is also well known that the models may suffer from missing parts and holes due to occlusion and insufficient scanning coverage of the scene. Furthermore, pointcloud or mesh representations of the scene may not be organized in a way that is intuitive to edit (i.e., they are unordered), and may take up large amounts of memory to store. These characteristics present a significant bottleneck for AR/VR applications to transmit customized user models in real time and to virtually interact with the models. 

To alleviate this problem, multiple lines of work such as \cite{tulsiani2018factoring, izadinia2017im2cad, li2015database,avetisyan2019scan2cad} have suggested that we decompose the scene into objects and their poses and scales. For example, one popular approach is to replace object pointcloud with a suitable Computer Aided Design (CAD) model from a database. The benefit of this is that the model is more likely to be complete and can be stored using its index in the database rather than a full mesh model. However, since models in CAD databases are most likely stored in a canonical pose and scale \cite{chang2015shapenet}, this approach is only useful when one can estimate the appropriate scale and pose together with recognizing the model category in the database. This leads to a \emph{joint shape retrieval and transform estimation} (JSRTE) problem, which is the focus of this paper.  

Another drawback of the above approach is that given a scene scan, there is no guarantee of finding similar models in the database. Thus we may lose a great deal of representation power if we only rely on referencing the database indexes. Nevertheless, it has been proposed that deep neural network (DNN) based 3D generative models such as \cite{achlioptas2017learning, park2019deepsdf, genova2019learning} allow one to produce a larger set of shapes than are available during training (or in the database) by learning a latent space that allows for interpolation and generalization of shapes. What this implies is that we may now be able to obtain the benefit of using a CAD model representation without sacrificing too much representation power.

In this work we utilize an SDF-based 3D DNN model to formulate the JSRTE problem constrained on similarity transform (rigid body transform and scale) as an optimization problem. We present a gradient-descent based solution and compare our results to the state of the art for shape retrieval and/or transform estimation problems. We show this approach produces excellent results on synthetic and real world data. Our contributions are as follows:
\begin{enumerate}
\vspace{-0.1cm}
    \item Formulate JSRTE problem as a novel joint optimization problem.
    \vspace{-0.1cm}
    \item Demonstrate a parameterization of the problem allowing for easy incorporation in popular DNN frameworks.
    \vspace{-0.1cm}
    \item Conduct experiments on synthetic and real datasets showing the effectiveness of our approach on JSRTE problem and as a form of 3D data compression.
\end{enumerate}
\vspace{-0.3cm}

\section{Related Work}\label{sec:related_works}
\vspace{-0.3cm}
\subsection{3D Shape Alignment using Signed Distance Functions}
\vspace{-0.2cm}
Aligning 3D shapes is a well studied problem with varying solutions depending on the representations of the 3D shape. For example, when the geometry of a 3D object is represented using pointcloud, solutions for estimating rigid-body and orientation-preserving similarity transformations have been proposed \cite{horn1987closed, horn1988closed, arun1987least, walker1991estimating,umeyama1991least}. 

Alternatively, and more pertinent to our work, 3D shapes can be aligned using signed distance functions (SDFs). \cite{Steiner2013StatisticalSM} provided a good overview of some of these methods. For example, \cite{slavcheva2016sdf,bylow2013real} presented techniques to estimate rigid-body transformations between two shapes represented using SDFs. \cite{mahmoodi2012similarity, cui2018robust} further tackled the problem of estimating rigid-body transformation and scale using SDFs, the same type of transformation studied in this paper. Their algorithms made use of geometric moments and the Fourier transform. In comparison, we develop a gradient-descent based algorithm to find a solution to an optimization formulation. Similarly, \cite{paragios2003non,huang2006shape} also made use of gradient-descent algorithms on optimization problems to estimate similarity transformations between SDFs. Yet, none of the works above explicitly allow for solving shape retrieval and transformation estimation in one optimization problem. Rather, they only deal with transformation estimation given the shape model.
\vspace{-0.2cm}
\subsection{3D Features for Shape Retrieval and Alignment}
\vspace{-0.2cm}
There exists a body of work that explicitly deals with shape retrieval. Classical shape retrieval algorithms such as \cite{shan2004linear, guo20143d, tombari2010object} usually follow a two-step approach: The first step involves shape recognition from a database by making use of shape descriptors such as \cite{tombari2010unique, rusu2009fast, aldoma2012our}. The second step involves a correspondence search using 3D keypoints and descriptors followed by registration. It has been argued in \cite{slavcheva2018signed} that dense alignment via SDFs can produce much better results since it does not make use of a correspondence search which may introduce errors. Our work favors this approach. We will compare our superior results against classical 3D feature based retrieval techniques.

More recent work has shifted from using hand-crafted descriptors to data driven ones such as \cite{zeng20173dmatch,deng20193d}. Alternative data driven approaches \cite{pham2018shrec} make use of deep neural networks (DNNs) to replace the recognition module entirely. However, the sequential approach of recognition and then registration does not allow for corrections of errors in the registration phase caused by a mismatch during the recognition stage. To address this problem, \cite{avetisyan2019end} proposed an end-to-end shape retrieval and alignment module. In comparison, although our work also jointly optimizes over the space of shapes and transforms, the formulation is not a feed-forward neural network but an optimization problem that makes use of a neural network model. The main benefit of this is that the optimization process can be viewed as introducing feedback that allows iterative minimization of errors from predictions made by the network. A feed-forward module can only make a prediction once per input and has no mechanism to minimize errors after training. More importantly, the space of shapes over which we  optimize allows us to retrieve objects that were not seen during training. 
\vspace{-0.2cm}
\section{DeepSDF-based Shape Retrieval and Similarity Transform Estimation}\label{sec:approach}
\vspace{-0.2cm}
\subsection{Problem Statement}\label{sec:problem_statement}
\vspace{-0.2cm}
Assume that we are given a set of shapes $\mathbb{S} =  \{ S_{1}, S_{2}, S_{3}, \hdots, S_{n} \}$ belonging to the same class, all in the canonical scale and pose (e.g., a set of chairs all normalized to lie in the unit sphere with forward directions aligned with the $+y$ axis). Further assume the shapes are represented as meshes although our approach also works for other representations as long as we can extract a signed distance function from the representation.

Then, for a new shape $S$ belonging to the same class but not necessarily in $\mathbb{S}$ or in the same canonical scale and pose, we would like to firstly build a dataset $\mathbb{D}$ of shapes using $\mathbb{S}$, and secondly find an item in $\mathbb{D}$ and a corresponding similarity transformation that best approximates $S$.

Note that we do not enforce $\mathbb{D}$ = $\mathbb{S}$, but only require that $\mathbb{D}$ is constructed using only $\mathbb{S}$. Typical shape retrieval problems tend to set $\mathbb{D}$ = $\mathbb{S}$. However, as we will show, this may restrict the richness of $\mathbb{D}$.

Our approach assumes that all given shapes have an associated SDF. This can be easily obtained, for example, from a triangle mesh representation of the shape.\footnote{Please refer to the supplementary for more details on SDF extraction and sampling.}

\begin{figure}[h!]
    \vspace{-0.2cm}
    \centering
    \includegraphics[width=0.8\textwidth]{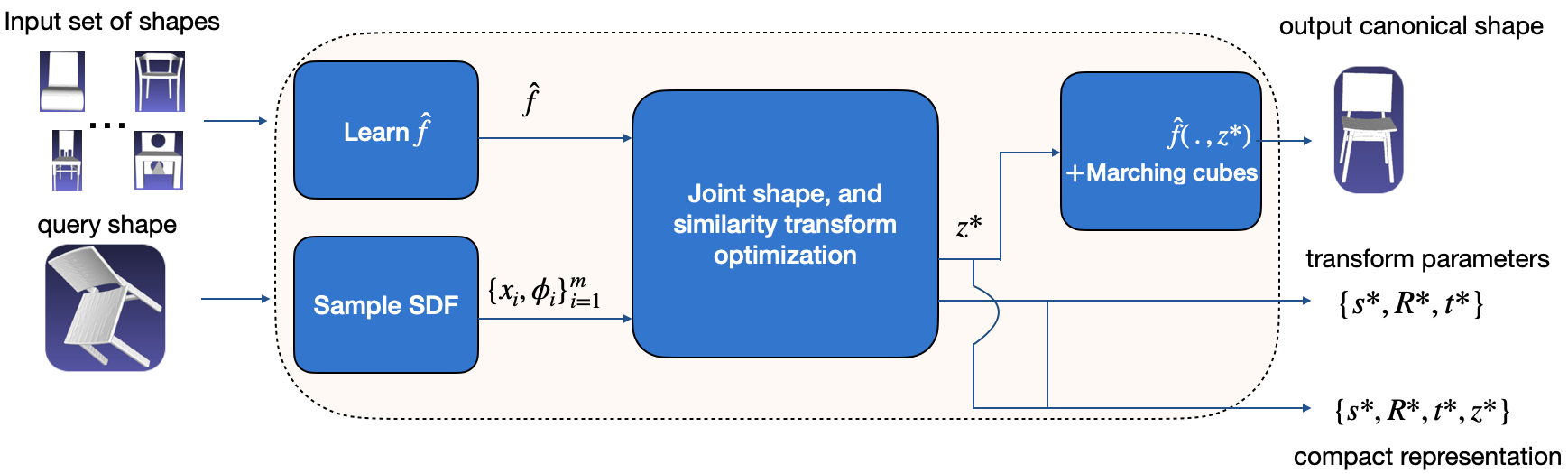}
    \vspace{-0.1cm}
    \caption{Overview of the proposed JSRTE algorithm. 
    }
    \label{fig:alg_img}
\end{figure}

\vspace{-0.2cm}
As shown in Fig.~\ref{fig:alg_img}, the input to our algorithm is a query shape $S$ and a set of canonical shapes $\mathbb{S}$.
Its output are the optimal shape representation in canonical pose and scale from $\mathbb{D}$ as a mesh, the optimal similarity transformation, and a compact representation of the query shape for data compression.
\vspace{-0.2cm}

\subsection{Notation and Preliminaries}\label{sec:notation_and_preliminaries}
\vspace{-0.2cm}
A 3D similarity transformation $g \in Sim(3)$ is a tuple $(s, R, t)$, with $s \in \mathbb{R}^{+}$, $R \in SO(3)$, $t \in \mathbb{R}^3$, and $R$ can be written as a $3\times 3$ real matrix. Then, $g$ acting on a point $x \in \mathbb{R}^3$ is given by
\begin{equation}\label{eq:similarity_def}
    g(x) = sRx + t.
\end{equation}

\vspace{-0.2cm}
Let $S$ be a solid 3D shape with a closed surface such that we have a well-defined notion of inside and outside. In addition, let $\partial \Omega_{S}^{\phantom{-}}$ represent the set of points on the surface of $S$, $\Omega_{S}^{\phantom{-}}$ represent the set of points on the surface and interior of $S$, and $\Omega_{S}^{c}$ the complement of $\Omega_{S}^{\phantom{-}}$. Then an SDF is an implicit representation of the shape defined as:
$
    \Phi(x, \Omega_{S}^{\phantom{-}}) = 
    \begin{cases}
        -\underset{y \in \partial \Omega_{S}^{\phantom{-}}}{\text{ min }} || x -y||_{2} \hspace{5pt}\text{,  if  } x \in \Omega_{S}^{\phantom{-}}; \\
        \underset{y \in \partial \Omega_{S}^{\phantom{-}}}{\text{ min }} || x -y||_{2} \hspace{5pt} \text{,  if  } x \in \Omega_{S}^{c}.
    \end{cases}
$

Intuitively, for a  fixed shape $S$, its SDF at a point $x$ tells us the distance from $x$ to the surface of $S$, with the sign determined by whether $x$ lies in the interior of $S$. 

Let $S, S^{'}$ be two shapes related by a similarity transformation $g$ as defined in \eqref{eq:similarity_def}, i.e., $\partial \Omega_{S^{'}} = \{g(x)| x \in \partial \Omega_{S}^{\phantom{-}}\}$.
Under this condition, it is well known \cite{paragios2003non} that
\begin{equation}\label{eq:sdf_similarity}
    \Phi(g(x), \Omega_{S^{'}}) = s\Phi(x, \Omega_{S}^{\phantom{-}}).
\end{equation}
In other words, SDFs are invariant to rotations and translations, but vary proportionally with isotropic scaling. The constant of proportionality is the scaling factor $s$.
\vspace{-0.2cm}
\subsection{Approach}
\vspace{-0.2cm}
Our proposed approach first learns a dataset $\mathbb{D}$ from the input set of shapes $\mathbb{S}$ and then formulates and solves an optimization problem to jointly estimate the transformation and shape parameters. Inspired by \cite{park2019deepsdf}, we also learn a generative model and latent space of shapes. The benefit to this is that the latent space is more expressive than the input set of shapes $\mathbb{S}$. Specifically, the latent space has been shown in \cite{park2019deepsdf} to extrapolate the shapes in $\mathbb{S}$ and produce new shapes of the same class but not contained in $\mathbb{S}$. The generative model also allows us to formulate searching over the latent space as a tractable optimization problem. In addition, since the dimension of the latent space is relatively small and a similarity transformation involves only a few parameters, we can simultaneously obtain a compact representation of the query shape. 

\subsubsection{Approximating SDFs with DNNs and Learning a Latent Dataset}
\vspace{-0.2cm}
Given an object class of shapes in a canonical pose and scale, $\mathbb{S} = \{ S_{1}, S_{2}, S_{3}, \hdots, S_{n} \}$, one can learn a generative model $\hat{f}(x,z) \approx \Phi(x, h(z))$ using DNNs \cite{park2019deepsdf}. Here $x \in \mathbb{R}^3$ is a point and $z \in \mathbb{R}^{d}$ is a latent vector with one-to-one correspondence with the shapes in $\mathbb{S}$. $h(z)$ is a mapping that assigns each latent vector in $\mathbb{R}^{d}$ to a shape in  $\mathbb{S}$, i.e., $ h: \mathbb{R}^{d} \rightarrow \mathcal{P}(\mathbb{R}^3)$ such that $\forall S \in \mathbb{S} \hspace{5pt} \exists z \in  \mathbb{R}^d : h(z) = \Omega_{S}^{\phantom{-}}$. The dimension $d$ is a design choice, and we refer to $\mathbb{R}^d$ as the latent space. It was also shown in \cite{park2019deepsdf} that in learning $\hat{f}$, one learns a latent space that may generate new shapes not in the training dataset. 

The implication of these results is that we can choose the latent space as the dataset for our algorithm. Also, given a shape $S$ in the same canonical scale and pose and a set of $m$ samples of its SDF $ \chi_{S}^{\phantom{-}}  = \{(x_{i}, \phi_{i} = \Phi(x_{i}, \Omega_{S}^{\phantom{-}})\}_{i=1}^{m}$, $S$ can be estimated by solving the following problem \cite{park2019deepsdf}: \vspace{-0.1cm}
\begin{equation}\label{eq:sdf_z_optimization}
    \underset{z \in \mathbb{R}^{d}}{\text{ min }} \sum_{(x_i, \phi_i) \in \chi_{S}^{\phantom{-}}}| \hat{f}(x_{i}, z) - \phi_{i} |.
\end{equation}

\vspace{-0.2cm}
 The above problem can be viewed as a special case of shape retrieval, since one can convert the latent vector back into a mesh by using $\hat{f}$ to generate SDF samples of the shape and using \cite{lorensen1987marching} to convert it to a mesh. Moreover, in the event that we have incomplete knowledge of the SDF of $S$, this formulation gives some robustness and has been shown to be useful for shape completion as well. 
 
 However, the above restriction to shapes in the canonical pose and scale limits the usefulness of this work in real world applications, since most objects are not in a canonical scale and pose. While it is possible to learn a new generative model on a larger dataset containing scaled and transformed versions of shapes in $\mathbb{S}$, we find that this is an inefficient approach to solving the problem. Our main contribution in this work is to present a principled way to overcome this problem without making use of any more data or computation for training.

\subsubsection{Joint Optimization of Shape and Transform Parameters}
\vspace{-0.2cm}
Our main idea is to modify \eqref{eq:sdf_z_optimization} to include an optimization over similarity transform parameters that is also easy to implement.
Using \eqref{eq:sdf_similarity} and \eqref{eq:sdf_z_optimization}, we formulate the problem as:
\begin{equation}\label{eq:sdf_z_trafo_1}
 \underset{s, R, t, z}{\text{ min }} \sum_{(x_i, \phi_i) \in \chi_{S}^{\phantom{-}}} \left| s \hat{f}\left( \frac{R^{-1}(x_i-t)}{s},z\right) - \phi_i \right|   
\end{equation}
with $ s \in \mathbb{R}^{+}, R \in SO(3), t \in \mathbb{R}^3, z \in \mathbb{R}^{d}$. 
Then one may use gradient-descent algorithms to solve the problem. However, care must be taken since $R$ does not live in a vector space. While it is possible to properly take gradient steps in $SO(3)$ \cite{dellaert2017factor}, we take an alternative approach to parameterize $R$. 

Using the axis-angle representation for rotation matrices, we know that for any rotation matrix $R$, there exists  $\omega: \left||\omega|\right|_{2} = 1$ and $ \theta \in [-\pi, \pi] :  R = \exp(\hat{\omega}\theta)$, where $\omega = [\omega_1, \omega_2, \omega_3]^{\top}$, $ 
    \hat{\omega} = \left[
    \begin{smallmatrix}
    0 & -\omega_3 & \omega_2 \\
    \omega_3 & 0 & - \omega_1 \\
    -\omega_2 & \omega_1 & 0
    \end{smallmatrix} \right].$
Since $\omega$ lies on the unit sphere, we may parameterize it using spherical coordinates as 
$
    \omega_1 = sin(\psi)cos(\rho), \hspace{5pt} 
    \omega_2 = sin(\psi)sin(\rho), 
    \hspace{5pt} 
    \omega_3 = cos(\psi) ,
$
where, to avoid potential confusion due to overloading symbols, we have used $\psi$ to represent the polar angle and $\rho$ to represent the azimuthal angle.
Consequently, we may write
\begin{equation}
    R(\psi, \rho, \theta) = 
    \exp\left(
    \begin{bmatrix}
    0 & -cos(\psi) & sin(\psi)sin(\rho) \\
   cos(\psi) & 0 & - sin(\psi)cos(\rho) \\
    -sin(\psi)sin(\rho) & sin(\psi)cos(\rho)& 0
    \end{bmatrix}
    \theta \right)
\end{equation}
and the initial problem \eqref{eq:sdf_z_trafo_1} as
 \begin{equation}\label{eq:sdf_z_trafo_2}
 \underset{ \substack{s, \psi, \rho, \\ \theta,t, z}}{\text{ min }} \sum_{(x_i, \phi_i) \in \chi_{S}^{\phantom{-}}} \left| s \hat{f}\left( \frac{[R(\psi, \rho, \theta)]^{-1}(x_i-t)}{s},z\right) - \phi_i \right|.
\end{equation}

\vspace{-0.2cm}
In this way, no special handling of gradients has to be taken. We find that this explicit formulation allows for easy implementation and incorporation into common deep learning frameworks with automatic differentiation such as \cite{NEURIPS2019_9015}. We summarize our JSRTE algorithm in Algorithm~\ref{alg:algo1}.
\vspace{-0.2cm}

\subsection{Implementation}\label{sec:implementation_details}
\vspace{-0.2cm}
We have implemented JSRTE algorithm using PyTorch \cite{NEURIPS2019_9015} and report some implementation details here for reproducibility. Specifically, the matrix exponential $\mbox{exp}(\hat{\omega} \theta)$ is expressed by Rodrigues' formula \cite{rodrigues1816attraction, ma2012invitation}:
$
    \exp(\hat{\omega} \theta) = I + \hat{\omega} sin(\theta) +  \hat{\omega}^2 (1- cos(\theta)).
$
Adam\cite{kingma2014adam} as implemented in \cite{NEURIPS2019_9015} is adopted to solve the optimization problem \eqref{eq:sdf_z_trafo_2}. To ensure that the search is in the feasible set of the problem, $ s \in [\underaccent{\bar}{s}, \bar{s}]$ can be restricted as $\underaccent{\bar}{s} = 0.01, \bar{s} = 10$. $\hat{f}$ is learned using open source code provided by \cite{park2019deepsdf, duan2020curriculum}, with latent vectors in $\mathbb{R}^{256}$. For all objects, we have trained on subsets of the corresponding Shapenet \cite{chang2015shapenet} class. To encourage convergence of the network, \cite{park2019deepsdf} has used a penalty term on the norm of $z$ weighed by $10^{-4}$. We also include this term in solving \eqref{eq:sdf_z_trafo_2}. We generally make use of the same SDF extraction and sampling approach used in \cite{park2019deepsdf} (see supplementary for detail).

\vspace{-0.2cm}
\begin{algorithm}[H]
\begin{algorithmic}[1]
\small
\Statex
\Function{JSRTE}{$\mathbb{S}, S, s^{0}, \rho^{0}, \psi^{0}, \theta^{0}, t^{0}, z^{0}$}
\Let{$\hat{f}, \mathbb{D}$}{$\mathbb{S}$} \Comment{Learn generative model and latent space from data (offline).}
\Let{$\{x_i, \phi_i\}_{i=1}^{m}$}{$S$} \Comment{Sample SDF from query shape}
\State Initialization \Comment{Initialize \eqref{eq:sdf_z_trafo_2} using provided parameters}
\State Solve \eqref{eq:sdf_z_trafo_2} for $s^*, R^*, t^*, z^*$ \Comment{Use gradient descent to solve \eqref{eq:sdf_z_trafo_2}}
\Let{$\{\bar{x}_i, \hat{f}(\bar{x}_i, z^*)\}_{i=1}^{n}$}{$\hat{f},z^*, \bar{x}_i \in [-1, 1]^3$} \Comment{Generate SDF samples at $z^*$ using $\hat{f}$}
\Let{$S^*$}{MARCHINGCUBES$(\{\bar{x}_i, \hat{f}(\bar{x}_i, z^*)\}_{i=1}^{n})$} \Comment{Generate mesh from SDF samples with \cite{lorensen1987marching}}
\State \Return{$s^*, R^*, t^*, z^*, S^*$}

\EndFunction
 \end{algorithmic}
 \caption{Joint Shape Retrieval and Transform Estimation (JSRTE)}
 \label{alg:algo1}
\end{algorithm} \vspace{-0.4cm}
\vspace{-0.2cm}
\section{Experiments}\label{sec:experiments}
\vspace{-0.3cm}
We perform three experiments to validate our approach for shape and transformation estimation. The first experiment seeks to evaluate the performance of the proposed algorithm when we initialize its parameters within reasonable bounds of the groundtruth. The second experiment evaluates our algorithm on real world data and compares it to state-of-the-art benchmarks. The third experiment assesses the viability of our approach as a means for 3D data compression.
\vspace{-0.2cm}
\subsection{Synthetic data}
\vspace{-0.2cm}
Gradient-descent based methods are susceptible to converging to locally optimal solutions. So it is important to find good initialization points for the variables. For this experiment, we want to observe the behavior of our algorithm when initialized within reasonable bounds of the groundtruth.

We sample objects from the chair, table, sofa, and bed categories from ShapeNet \cite{chang2015shapenet}. First, 30 objects from each category are sampled. Then for each object a random scale and pose are assigned by sampling the groundtruth parameters, $s \sim \mbox{Uniform}([\frac{1}{2}, 2) )$ , $ \psi, \rho, \theta, \sim \mbox{Uniform}([-\pi, \pi))$, and $||t||_2 \sim \mbox{Uniform}([0, 4)) $. The direction of $t$ is chosen uniformly from the surface of a unit sphere. For each shape and transform pair, we solve the shape retrieval problem using our algorithm 50 times, with each trial using a random initialization. 
We carry out two sub-experiments to highlight scenarios where our algorithm is applicable. In the first scenario, we assume that we know the axis of rotation (and consequently $\psi, \rho$), but all other parameters are unknown. This is a reasonable assumption since many objects lie upright on flat surfaces such as floors or tables. One may use the normal to the surface as the axis of rotation. In the second scenario, we relax this assumption and introduce some uncertainty in the axis of rotation. The scenarios are used to randomly sample the initialization for our algorithm.

Concretely, we refer to the initialization parameters as $s^{0} = s(1 + \Delta s), \psi^{0} = \psi + \Delta \psi, \rho^{0} = \rho + \Delta \rho, \theta^{0} = \theta + \Delta \theta, t^{0} = t + \Delta t$. The $\Delta $ variables correspond to the difference between the ground-truth and initialization parameters. We then sample the $\Delta $ variables uniformly from ranges determined by each scenario. Table~\ref{table:shapenet_exp_0} below summarizes these ranges for the different scenarios.
\begingroup
\renewcommand{\arraystretch}{1.5}
    \begin{table}[!ht]
    \small
    \centering
    \vspace{-0.3cm}
    \caption{Parameter ranges for the difference between initialization points and groundtruth values.}
        \begin{tabular}{c c c c c c}
 \hline
        Scenario &   $\Delta s$ & $\Delta \psi$ & $\Delta \rho $ & $\Delta \theta$ & $ ||\Delta t||(m) $ \\
        \hline
        \text{Known rotation axis } & $[0, 0.3)$ & N/A & N/A & $[-\frac{\pi}{9}, \frac{\pi}{9})$ & $[0, 0.15)$ \\
        \hline
        \text{Unknown rotation axis } & $[0, 0.3)$ & $[-\frac{\pi}{36}, \frac{\pi}{36})$ & $[-\frac{\pi}{36}, \frac{\pi}{36})$ & $[-\frac{\pi}{9}, \frac{\pi}{9})$ & $[0, 0.15)$ \\
        \bottomrule
        \end{tabular}
        \label{table:shapenet_exp_0}
    \end{table}
\endgroup

\vspace{-0.3cm}
In all scenarios, We choose the direction of $\Delta t$  uniformly from the surface of a unit sphere. $z^{0} \sim \mbox{Normal}( \mathbf{0} , \sigma \mathbf{I}), \sigma = 0.01$. We use the corresponding ShapeNet \cite{chang2015shapenet} model class as input shape set $\mathbb{S}$.

To quantitatively measure the performance of our algorithm, we use the F-score between the output shape (transformed by our predicted transform parameters) and the test shape as recommended in \cite{tatarchenko2019single}. The F-score measures the similarity between 3D surfaces as the harmonic mean between precision and recall. A point on the predicted surface is considered a true positive if it is within a threshold $\epsilon$ of the groundtruth surface. In this work, we have set this threshold as $5\%$ of the scale of the groundtruth object. We refer to this score as F@$5\%$. The F-score ranges from $0$ to $1$, with higher numbers indicating more closely matched shapes.

\begin{figure*}[!ht]
    \centering
    \vspace{-0.2cm}
    \begin{subfigure}{0.4\textwidth}
    \includegraphics[width=\linewidth]{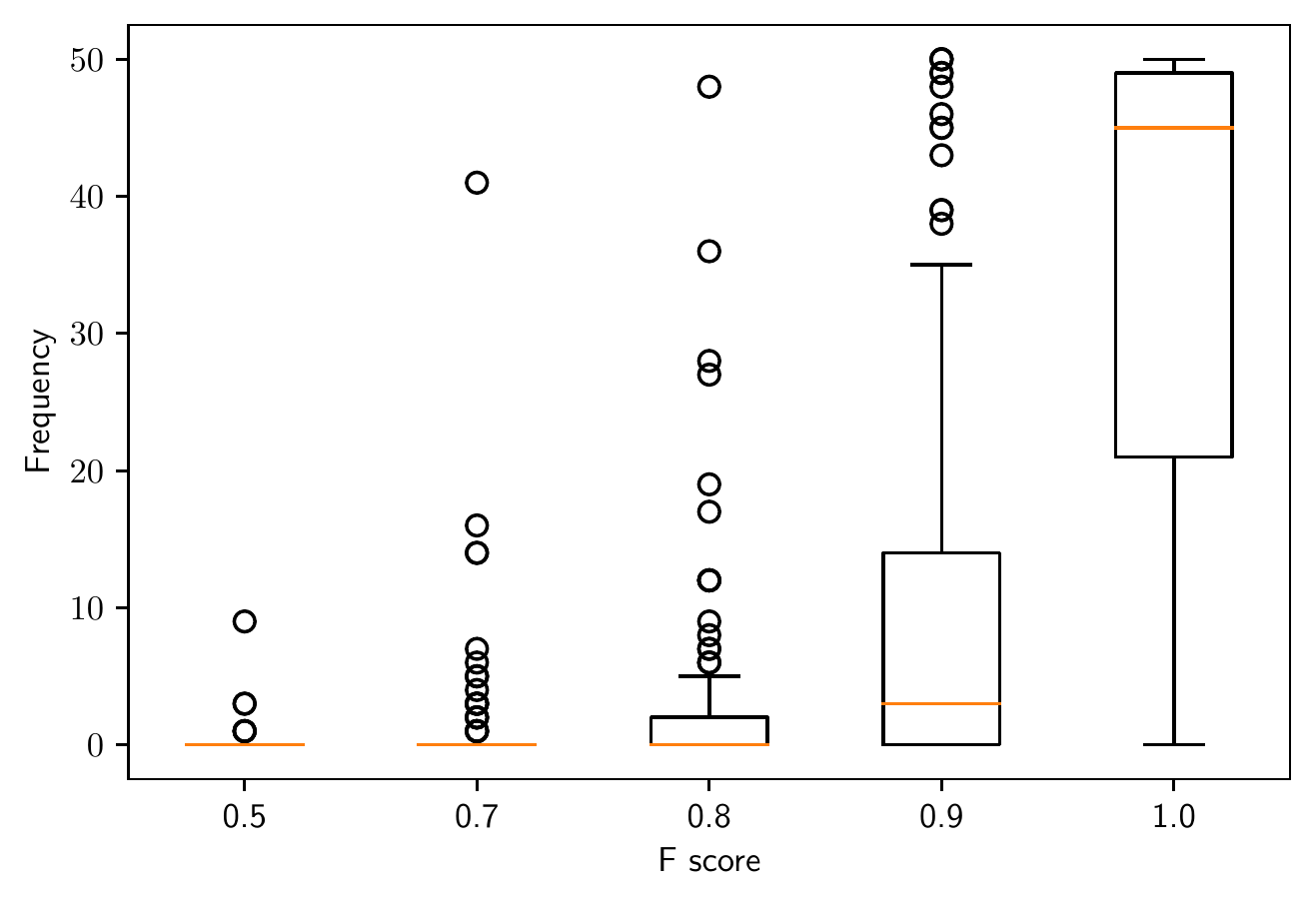} 
    \end{subfigure}%
    \begin{subfigure}{0.4\textwidth}
    \includegraphics[width=\linewidth]{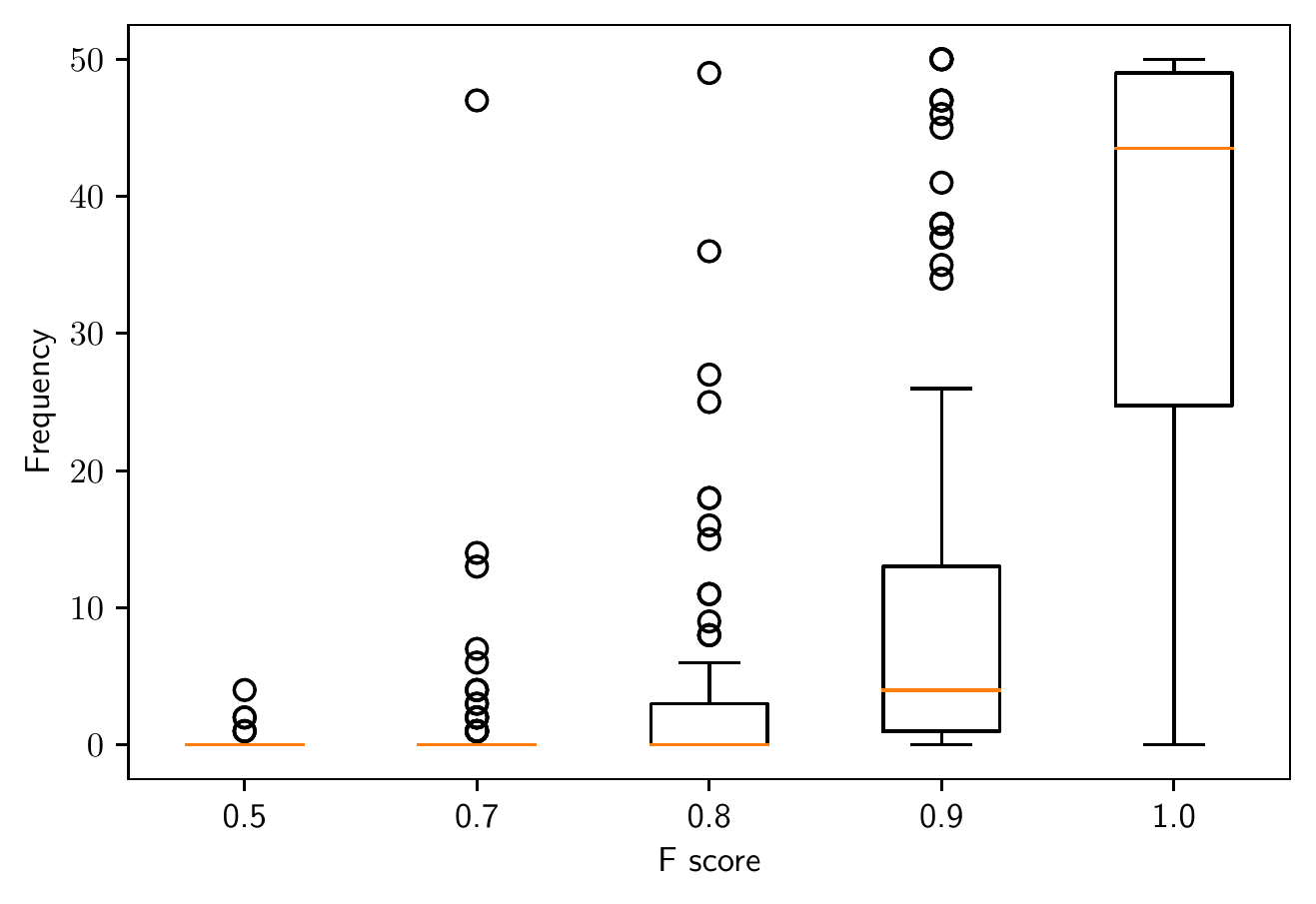}
    \end{subfigure}%
    \vspace{-0.1cm}
    \caption{Box plot of F@$5\%$ from experiments described in Table~\ref{table:shapenet_exp_0}. \textbf{Left:} Scenario with a known axis of rotation; \textbf{Right:} Scenario with an unknown axis of rotation. The median for each bin is shown as an orange line in the box, with outliers shown as circles. Each box is placed on the right edge of its corresponding bin, the left edge of each bin is the right edge of the preceding bin.}
    \label{fig:f_score_all}
\end{figure*}

For each shape and transform pair, we categorize the F-score into bins and count the number of initializations out of 50 that fall into each bin. 
Then for each bin, we report aggregate statistics using box plots in Fig.~\ref{fig:f_score_all}. The position of each box-plot on the x-axis corresponds to the rightmost edge of its bin. The leftmost edge of its bin is the previous point on the x-axis.

\vspace{-0.3cm}
Fig.~\ref{fig:f_score_all} corresponds to scenarios with a known and unknown axis of rotation respectively. In both scenarios, we observe that our algorithm performs well with most F@$5\%$ scores greater than 0.8. We also observe a slight degradation in performance in the scenario with an unknown axis of rotation, as is to be expected due to more uncertainty. Intuitively, the results imply that with a known axis of rotation, if we know the amount of rotation up to $\pm 20^{\circ}$, the translation up to $\pm 15$ cm and the scale up to $30\%$ of the true scale, we can expect our algorithm to perform very well. We also perform more experiments to test the our algorithm in other parameter ranges and report those as well as qualitative results in the supplementary.
\vspace{-0.2cm}
\subsection{Real Data}
\vspace{-0.2cm}
In the next set of experiments, we evaluate our algorithm on real data. In this paper, we demonstrate the results on the chair subset of the Redwood dataset \cite{choi2016large}. The Redwood dataset contains scans (represented as triangle meshes) of real objects in varying scales and poses. We have manually segmented the dataset so that we remove other objects in the background, leaving behind only the floor and the main object in each scene. We assume that all objects in the scene lie on the floor and are upright. This allows us to obtain an estimate for $\omega$ as the normal to the floor plane. In this way, we collect 89 shapes and samples of their SDFs. However, we use 31 randomly selected shapes from these 89 for validating the baselines described below, leaving only 58 for testing. We discarded all other scenes that could not be segmented properly. For each shape, we initialize $s$ using the radius of its bounding sphere and $t$ using the center of the bounding sphere. We estimate $\omega$ using the normal vector to the floor plane and $z$ is initialized by drawing from $\mbox{Normal}( \mathbf{0} , \sigma \mathbf{I}), \sigma = 0.01$. The only parameter left is $\theta$. For $\theta$ we grid up the space in increments of $30^\circ$ and run our algorithm for each value of $\theta^{0} \in [0, \frac{\pi}{6}, \frac{\pi}{12}, \hdots 2 \pi)$. We then select the best result from each of these initialization of $\theta$ using the F@$5\%$ score.

We compare our work to three different approaches for shape retrieval and transform estimation. In these approaches, we perform recognition to find the closest shape to the query shape in the set of input shapes and then perform registration to estimate a similarity transform.\footnote{ We are aware of other state-of-the-art SDF based methods for jointly estimating shape and similarity transformation \cite{avetisyan2019end}. However, we could not find a publicly available implementation to use for testing.}

The first approach uses Harris3D keypoints and the SHOT\cite{salti2014shot} descriptor to perform recognition and provide correspondences for registration. Registration is then carried out using \cite{umeyama1991least}. Specifically, for recognition, we describe each shape in the dataset using the descriptor and keypoint detecting method stated above. During test time, we compare the keypoints and descriptors of the query shape to those in the dataset. This method is referred to as Harris3D+SHOT.

The second approach replaces the keypoint detector and descriptor with the shape descriptor developed in \cite{aldoma2012our}. For registration, points are sampled from the mesh and then applied to an Iterative Closest Point (ICP)\cite{besl1992method} variant that makes use of \cite{umeyama1991least} for transform estimation. Correspondences are assigned by solving an optimal assignment problem. We find that this significantly improves the result. This method is referred to as OURCVFH+ICP. 

The last approach uses PointNet\cite{qi2017pointnet} as described in \cite{pham2018shrec} for recognition and the ICP method described above for registration. This method is referred to as PointNet+ICP.

\begin{figure}[!ht]
\setlength\tabcolsep{2pt}
\centering
\begin{tabular}{ccccc}
 \textbf{Test Shape} &
 \textbf{Harris3D + SHOT} &
 \textbf{OURCVFH + ICP} &
 \textbf{Pointnet + ICP} &
 \textbf{Ours}\\
 \includegraphics[width=0.20\textwidth]{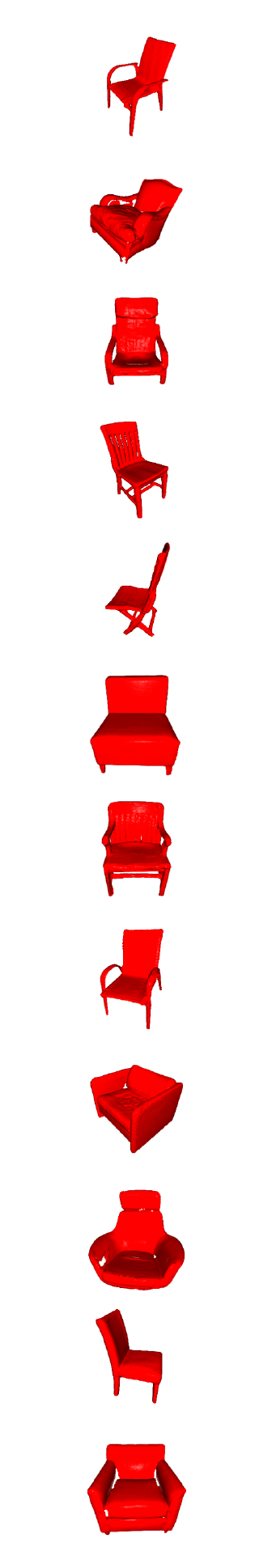}  &
 \includegraphics[width=0.20\textwidth]{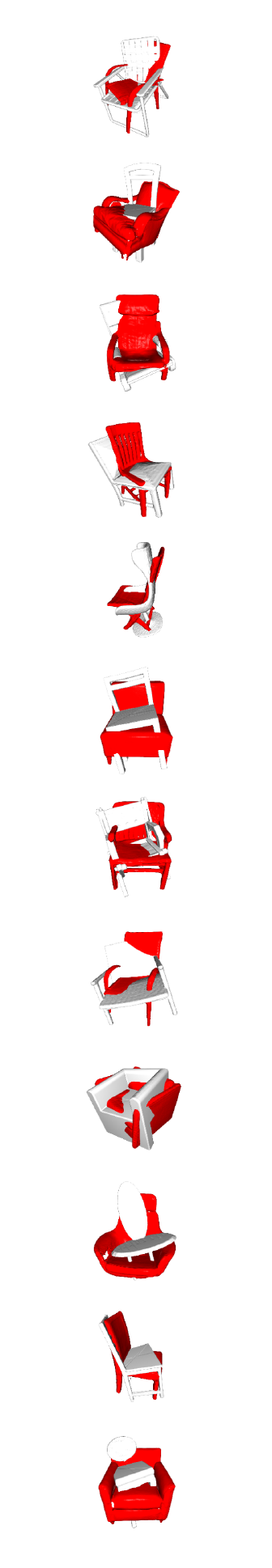} &
 \includegraphics[width=0.20\textwidth]{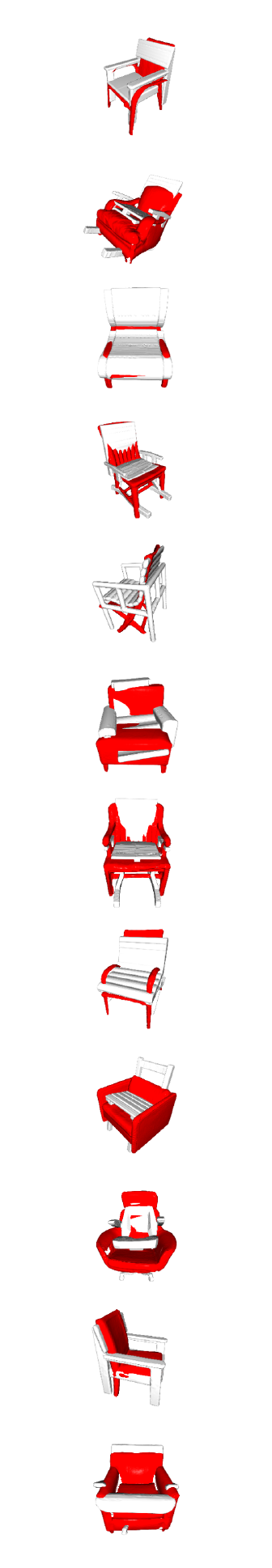} &
 \includegraphics[width=0.20\textwidth]{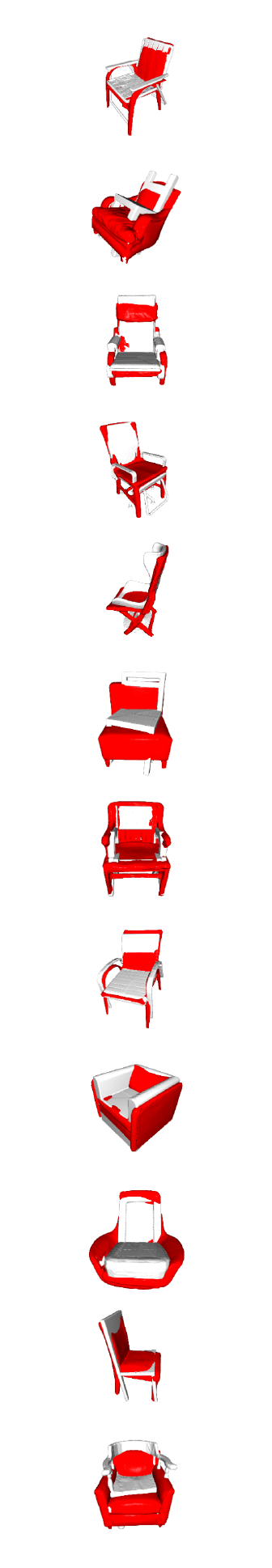}  & 
  \includegraphics[width=0.20\textwidth]{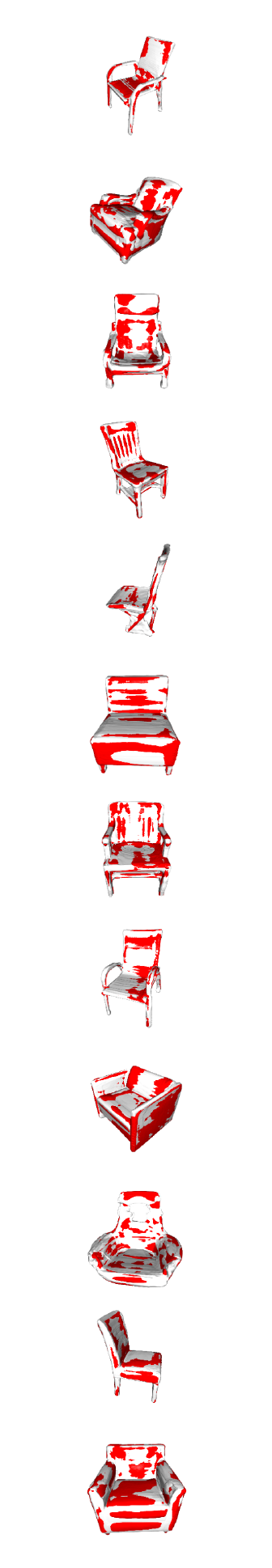} \\
\end{tabular}
\vspace{-0.2cm}
\caption{Qualitative result on Redwood dataset. Left to Right, test shape, results using Harris3D and SHOT\cite{salti2014shot} for shape retrieval, results using OURCVFH \cite{aldoma2012our} and ICP, results using PointNet \cite{qi2017pointnet} and ICP, our proposed method. We have shown each method's result in gray overlaid with the input mesh in red. Our algorithm provides significant qualitative improvement with almost perfect retrieval.}
\label{fig:redwood_1}
\end{figure}

We report the average F@$5\%$ score of these methods in Table~\ref{table:redwood_results}, which shows our algorithm significantly outperforms the three alternative approaches. We also provide qualitative results of randomly sampled test cases in Fig.~\ref{fig:redwood_1}, which shows that the shapes predicted by our method are qualitatively better than the alternative methods, with much less misalignment. Alternative methods using a two-step pipeline of recognition and registration can propagate errors in the recognition stage that cannot be corrected by just a similarity transformation in the registration stage. In addition, our method uses a richer search space for shapes by using the latent space as its dataset, rather than just the set of input shapes provided. Moreover, finding good correspondences can be difficult for other two-step approaches as clean CAD models and noisy real scans have differing local surface properties. Finally, the need to sample the surface to allow tractable registration may introduce errors in the sense that true corresponding pairs may not belong to the sampled sets. Our method on the other hand does not make explicit use of correspondences.
\vspace{-0.2cm}
\begin{table}[!ht]
    \vspace*{-0.4cm}
    \caption{Quantitative results on Redwood dataset. Average F-scores for shape retrieval and similarity transformation methods compared to ours. Our algorithm's score more than doubles the others.}
    \label{table:redwood_results}
    \centering
    \begin{tabular}{ c c c c c }
        \toprule
         &  Harris3D + SHOT & OURCVFH + ICP & PointNet + ICP &  Ours\\ 
         \cmidrule{2-5}
     F@$5\%$    &  0.273 & 0.367  & 0.452 & \textbf{0.902} \\
     \bottomrule
    \end{tabular}
    \vspace*{-0.27cm}
\end{table}

\subsection{3D Shape Compression}
\vspace{-0.2cm}
Finally, we conduct an experiment to explore the viability of our approach as a form of 3D data compression. Given the neural network parameters of $\hat{f}$, we can choose to store only the latent vector representing a shape and the associated similarity transform. With access to $\hat{f}$ and the transform parameters, one can then decode the latent vector and convert it back to a mesh model. In our experiment, the network size is approximately 7MB. This is a constant cost shared across many uses and is consequently amortized.

For this experiment, we save each mesh in our Redwood test set under two mesh simplification schemes, mesh simplification using Quadric Error Metrics\cite{garland1997surface,Zhou2018} and Vertex Clustering \cite{Zhou2018}. We have varied the parameters of each method. For mesh simplification using Quadric Error Metrics, we set the target number of faces to be reduced by a factor of 100 and 1000 from the number of faces in the original mesh. We refer to these experiments as QE-100, QE-1000 respectively. For Vertex Clustering, we vary the size of the grid cells used for clustering to be a factor of 0.1 and 0.2 of the radius of the bounding sphere for the shape. We refer to these as VC-0.1, VC-0.2 respectively. We save all meshes and compute the average size of the files as well as the mean F-score (compared to the original mesh) to give an indication of the representation power at each simplification parameter. We compare these scores against our mean F-score and the cost of saving the latent vector and transformation parameters as a $4\times 4$ matrix and report the results in Table~\ref{table:redwood_compression_results}. The result shows that our method provides an impressive tradeoff between reconstruction quality and storage size.

\begin{table}[!ht]
    \vspace{-0.4cm}
    \caption{Comparison of storage size and representation power for various mesh simplification algorithms. Our approach provides an excellent trade-off between accuracy and storage space, saving over 20x space as the most accurate method with less than a 10\% drop in accuracy.}
    \label{table:redwood_compression_results}
    \centering
    \begin{tabular}{ c c c c c c }
         \toprule
         &  Ours & QE-100 & QE-1000 & VC-0.1 & VC-0.2\\ 
         \midrule
     Average F-score    &  0.902 & 0.977 & 0.778 & \textbf{0.978} & 0.839 \\
     \midrule
     Average size (KB) & \textbf{1.088} & 20.120 & 3.454 & 34.833 & 9.073 \\
     \bottomrule
    \end{tabular}
\end{table}

\vspace{-0.2cm}
\section{Conclusion}\label{sec:conclusion}
\vspace{-0.3cm}
In this paper we have presented a formulation for shape and similarity transform estimation from SDFs as an optimization problem. We have obtained good results with a gradient-descent optimization scheme and good initialization of the parameters. We have also shown results on a dataset of real scans and obtained superior performance to benchmarks. Our method currently relies on good segmentation of the scene and knowledge of the class of objects. In addition, it may struggle to represent fine details and may sometimes lead to disconnected meshes. As future work, we plan to tackle these problems as well as incorporating this method as a tool for 3D indoor scene understanding and representation. 
\section*{Broader Impact}
The work presented in the paper has the potential to be applied to a wide range of Computer Vision tasks. We focus on concrete application domains of Mixed Reality and 3D scene understanding. 

We envision that our work can be used as a part of Mixed Reality applications that require automatic content generation from real world scenes. Such applications include telepresence and automatic 3D modeling. While these applications are yet to reach full maturity we believe our work can contribute to the development of these applications. In addition, when coupled with other tools to help predict what is in a scene, our work can be used to help answer questions regarding "where is the object in the scene and how can we manipulate it?". These questions lie in the domain of 3D scene understanding with potential use cases in robotics. 

On the other hand, mixed reality applications require privacy preserving algorithms. As discussed in \cite{speciale2019privacy}, and illustrated in \cite{pittaluga2019revealing}, Mixed Reality applications may require cloud services for data storage, transmission and processing. Adversarial attacks on cloud infrastructure could then prove catastrophic since mixed reality applications tend to take place in private spaces such as homes. While the work described in this paper is amenable to cloud applications due to its compression properties, it is not naturally privacy preserving. Adversarial access to neural network parameters could reveal information about the 3D geometry and spatial  content of private spaces. Hence, extra measures such as the use of encryption must be put in place to protect privacy. Likewise, applications, such as robotic manipulation tasks, that may depend on our method for estimating the pose of objects may need to put measures in place to mitigate catastrophic results in failure cases. Wrong pose estimation could lead to damaging object while trying to manipulate them or even fatal injuries to humans in the vicinity.

Taking all this into account, we see the opportunity for more research on ways to apply shape retrieval methods such as our's in Mixed Reality and robotic applications while preserving privacy and detecting failure modes. We hope that this work encourages more discussion among researcher, engineers and consumers on the impact of such technologies and ways to mitigate negative consequences.
\section*{Acknowledgement}
This work is supported in part by the Office of Naval Research (ONR) under grant N00014-19-1-2055.
\newpage
\section*{Supplementary}
\addcontentsline{toc}{section}{Supplementary}
\renewcommand{\thesection}{\Alph{section}}
\setcounter{section}{0}
\section{Overview}
We provide additional quantitative and qualitative results on the algorithm presented in the main paper. We also provide more implementation details and descriptions of the metrics used. 

\section{Metrics}\label{sec:metrics}
We have made use of the F-score as recommended in \cite{tatarchenko2019single} to provide a quantitative evaluation of our work. Given a groundtruth shape $S$ and an estimate shape $\hat{S}$, as well as samples of points on their surfaces $\partial \Omega_{S}^{\phantom{-}}$ and $\partial \Omega_{\hat{S}}^{\phantom{-}} $, \footnote{For simplicity of notation, we have retained the same notation as that used for the set of all points on the surface. To compute the F-score, we can only use a sampling of the set of points on the surface.} the F-score is the harmonic mean of precision and recall metrics. Precision and recall are defined as follows: 

To compute the precision, for each point $y$ in the estimate shape, we first compute its distance to the groundtruth as:
\begin{equation*}
    e_{y} = \underset{x \in \partial \Omega_{S}^{\phantom{-}}}{\text{min}} || y - x||_2.
\end{equation*}

The precision is then computed as the percentage of points within distances less than $\epsilon_{p}$:
\begin{equation}
    P(\epsilon_p) = \frac{100}{|\partial \Omega_{\hat{S}}^{\phantom{-}}|} \sum_{y \in \partial \Omega_{\hat{S}}^{\phantom{-}}} \mathbf{1}_{\{e_{y} < \epsilon_{p}\}}.
\end{equation}

Recall is computed in the opposite direction.  Again, the error is first computed  for each point $x$ on the groundtruth shape as:
\begin{equation}
    e_{x} = \underset{y \in \partial \Omega_{\hat{S}}^{\phantom{-}}}{\text{min}} || x - y||_2,
\end{equation}
and recall as:
\begin{equation}
     R(\epsilon_r) = \frac{100}{|\partial \Omega_{S}^{\phantom{-}}|}\sum_{x \in \partial \Omega_{S}^{\phantom{-}}} \mathbf{1}_{\{e_{x} < \epsilon_{r}\}}.
\end{equation}

Finally, the F-score is computed as:
\begin{equation}
    F(\epsilon) = \frac{2P(\epsilon)R(\epsilon)}{P(\epsilon) + R(\epsilon)}
\end{equation} 
The F-score ranges from 0 to 1, with higher values indicating a better match between the estimate and the groundtruth. The F-score is very sensitive to deviations in shape and thus makes an excellent metric. To obtain pointclouds from the meshes, we have sampled 3000 points from the surface of each mesh. We have also set $\epsilon = 0.05r$, where $r$ is the radius of the bounding sphere of the shape $S$. 

An alternative metric may have been computing the distance between the estimated parameters and the groundtruth parameters, however, we find that solutions that are close in 3D space may not be close in the parameter space. For example, we find that the learned shapes can have different centers from their groundtruth counterparts, as such a good result may have a different translation value than the groundtruth. 

\section{ Additional Results}\label{sec:supp_experiments}

In this section, we provide additional results to support those presented in the main paper. 
\subsection{Synthetic data}
We perform additional experiments making use of synthetic data to observe the behavior of our algorithm under varying initialization conditions. Following the same procedure described in Section 4.1 of the main paper, we conduct two additional experiments that vary the initialization parameters as follows:
\begingroup
\renewcommand{\arraystretch}{1.5}
    \begin{table}[!ht]
    \small
    \centering
    \vspace{-0.3cm}
    \caption{Parameter ranges for the difference between initialization points and groundtruth values.}
        \begin{tabular}{c c c c c c}
 \hline
        Scenario &   $\Delta s$ & $\Delta \psi$ & $\Delta \rho $ & $\Delta \theta$ & $ ||\Delta t||(m) $ \\
        \hline
        \text{Known rotation axis } & $[0, 0.5)$ & N/A & N/A & $[-\frac{2\pi}{9}, \frac{2\pi}{9})$ & $[0, 0.20)$ \\
        \hline
        \text{Unknown rotation axis } & $[0, 0.5)$ & $[-\frac{\pi}{36}, \frac{\pi}{36})$ & $[-\frac{\pi}{36}, \frac{\pi}{36})$ & $[-\frac{2\pi}{9}, \frac{2\pi}{9})$ & $[0, 0.20)$ \\
        \bottomrule
        \end{tabular}
        \label{table:supp_shapenet_exp_0}
    \end{table}
\endgroup

In the first scenario in Table~\ref{table:supp_shapenet_exp_0},  we assume a known axis a rotation, but increase the range of values that the initialization parameters can take on compared to those presented in the main paper. In the second scenario, we relax the assumption of a known axis of rotation and introduce some uncertainty. We retain the same increase in the ranges of the other parameters. The purpose of this experiment is to see how the performance degrades when we introduce more uncertainty. 

\begin{figure*}[!ht]
    \centering
    \vspace{-0.2cm}
    \begin{subfigure}{0.45\textwidth}
    \includegraphics[width=\linewidth]{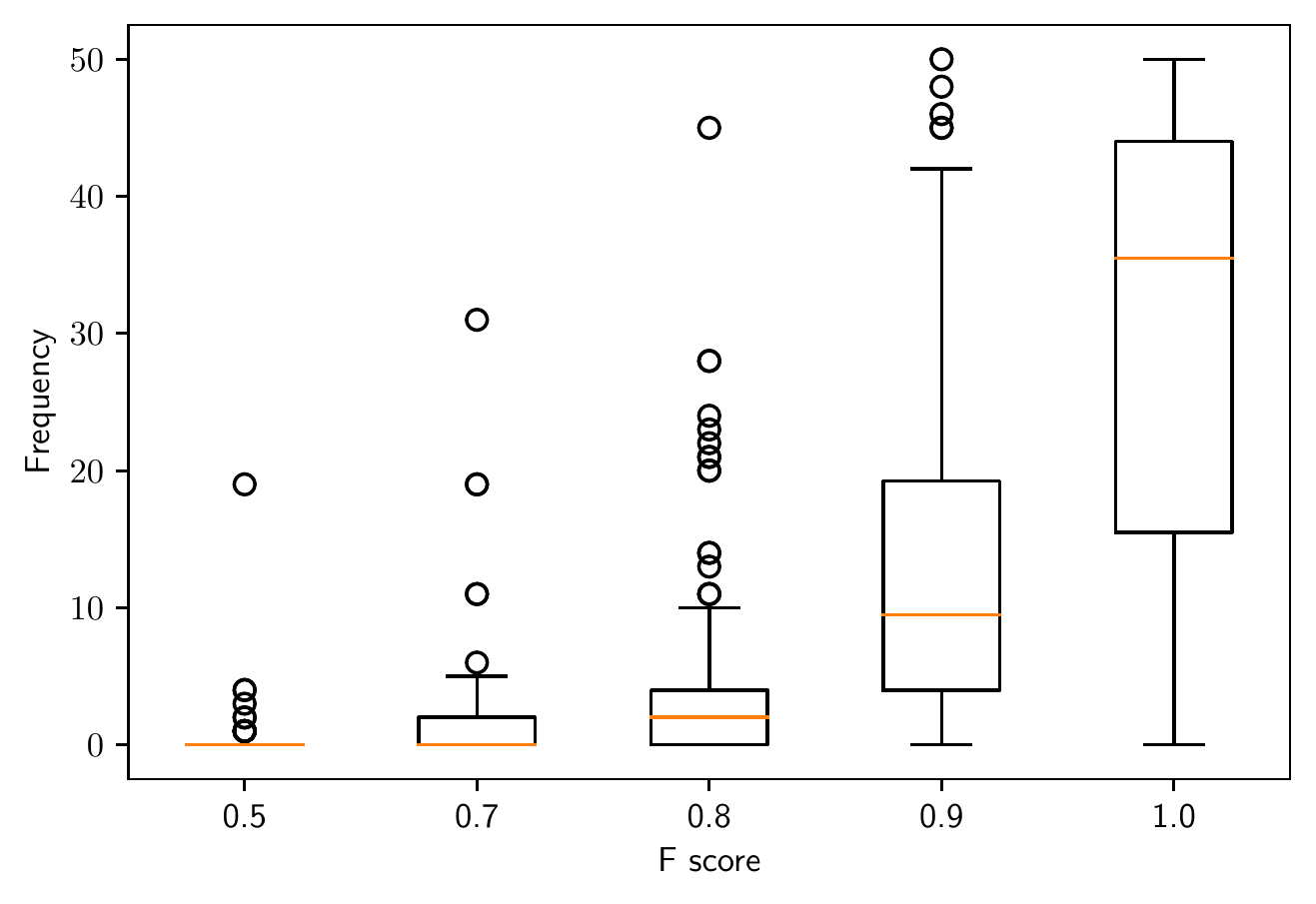} 
    \end{subfigure}%
    \begin{subfigure}{0.45\textwidth}
    \includegraphics[width=\linewidth]{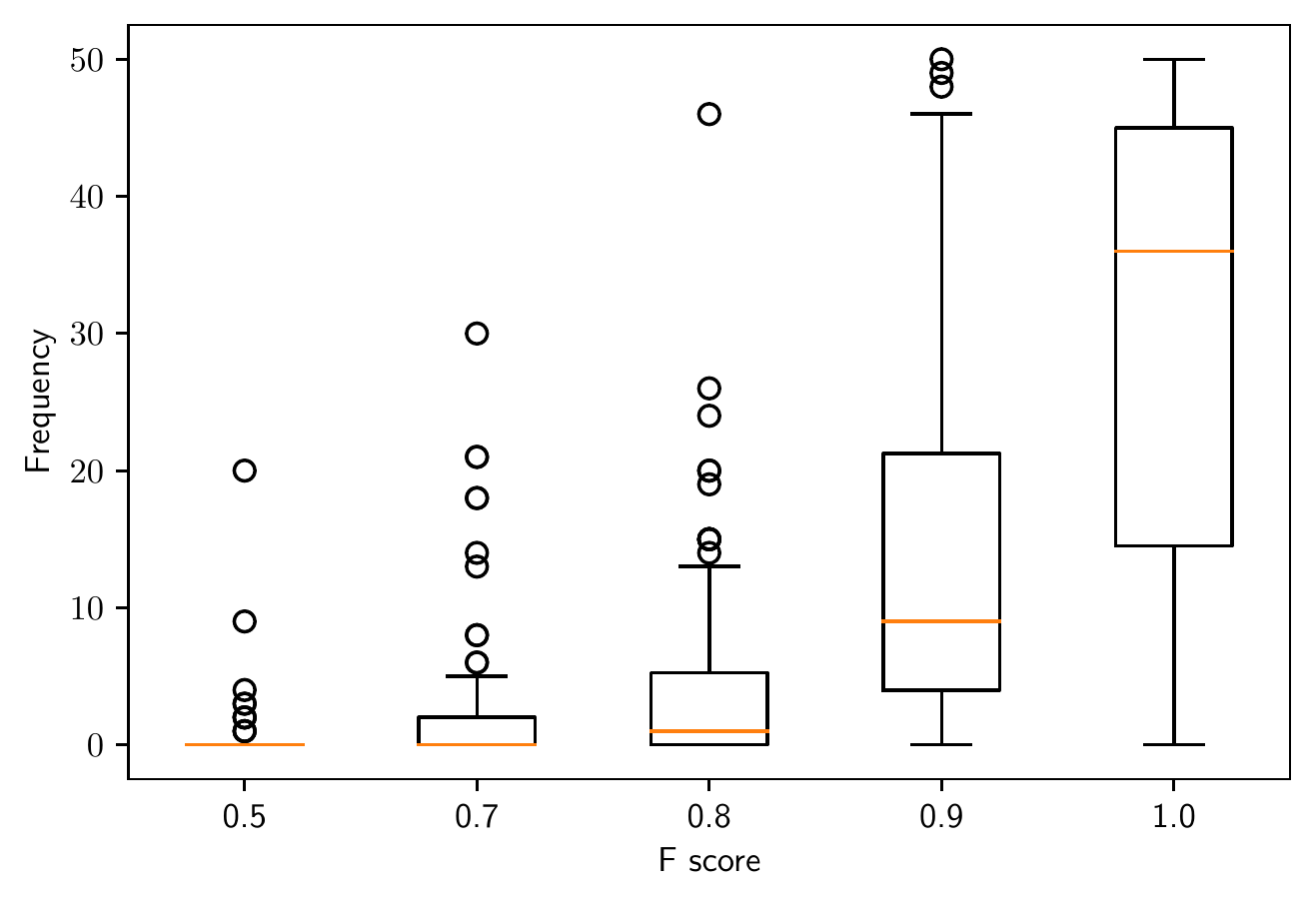}
    \end{subfigure}%
    \vspace{-0.1cm}
    \caption{Box plot of F@$5\%$ from experiments described in Table~\ref{table:supp_shapenet_exp_0}. \textbf{Left:} Scenario with a known axis of rotation; \textbf{Right:} Scenario with an unknown axis of rotation. The median for each bin is shown as an orange line in the box, with outliers shown as circles. Each box is placed on the right edge of its corresponding bin, the left edge of each bin is the right edge of the preceding bin. We still observe most median F-scores greater than 0.80, although we now see more cases of F-scores in the range $[0.5, 0.7]$. }
    \label{fig:f_score_all_2}
\end{figure*}

As shown in Fig.~\ref{fig:f_score_all_2}, we observe that both scenarios perform worse than those presented in the main paper. We also observe that for most of the experiments the median F-score is still greater than 0.8. The difference here is that we observe significantly more F-scores in the $[0.5, 0.7]$ range. 

\subsection{Real Data}
Figures~\ref{fig:supp_redwood_1} -~\ref{fig:supp_redwood_6} provide qualitative results from all the experiments carried out on the Redwood dataset \cite{choi2016large} detailed in Section 4.2 of the main paper. We have displayed comparisons to the benchmark making use of PointNet \cite{qi2017pointnet} and a variant of ICP \cite{besl1992method} described in the main paper. We report a similar level of performance as that presented in the main paper. Our algorithm consistently provides qualitatively better results with fewer shape mismatch and misalignments. The figures also indicate that our algorithm sometimes struggles with fine details and star shaped legs. This is partly due to poor convergence of the SDF model during training and also because in reality the star-shaped legs are free to rotate separately from the main body of the chair. These additional degrees of freedom are not well captured by our assumptions. 

\section{Implementation Details}
\subsection{SDF Sampling}
For experiments using synthetic data (Section 4.1 of the main paper), we follow the same sampling procedure described in \cite{park2019deepsdf}. For experiments using the Redwood dataset \cite{choi2016large} (Section 4.2 of the main paper), we first preprocess the data by manually segmenting the floor and main object in each scene. We then make use of  the plane fitting algorithm provided by \cite{Zhou2018} to segment the main object and the floor. We also obtain the axis of rotation in this fashion as the normal to the floor plane. Furthermore, we make use of the hidden point removal algorithm described in \cite{katz2007direct} and implemented in \cite{Zhou2018} to remove artifacts. Other artifacts and noisy points are also removed using the statistical outlier removal algorithm provided by \cite{Zhou2018}. 

Finally, for each point in the preprocessed mesh, we sample new points at a distance of $\pm 0.01$m in the direction of the normal to the surface at each point. We estimate the SDF at these new points as $\pm 0.01$ respectively. To obtain SDF samples representing the freespace, we take each point on the mesh and randomly and uniformly sample a new point between $0.07$m and $0.20$m away from the point on the mesh, in the direction of the outward pointing normal. We compute the SDF at these new points by finding the approximate distance to the closest point on the surface of the mesh. The sign is then determined by the whether each new point is in the direction of the outward pointing normal at its nearest neighbor on the mesh. We use a k-d tree and the set of vertices on the densely sampled meshes to approximate the nearest neighbors. We obtain 25000 freespace SDF samples in this manner. 

The SDF samples used during optimization are composed of the freespace SDF as well as the SDF sampled at distances $\pm 0.01$m  from points on the mesh.

\subsection{Optimization Parameters}
To solve the proposed optimization problem, we employ Adam \cite{kingma2014adam} as implemented in \cite{NEURIPS2019_9015}. We follow a stochastic gradient-descent approach and make use of 8000 SDF samples at each iteration. The learning rate is set to $0.05$ for all parameters except for those assumed to be known. The learning rate is decreased by a factor of $5$ every 400 iterations. We solve the optimization problem for 800 iterations. 

\subsection{Real data benchmarks}
We report some implementation details for the benchmarks used for experiments on the Redwood dataset \cite{choi2016large} presented in the main paper. 

\subsubsection{Harris3D + SHOT}
For this benchmark we make use of Harris3D keypoint detector and the SHOT\cite{tombari2010unique} descriptor as implemented in \cite{rusu20113d}. 

\subsubsection{OURCVFH + ICP}
For this benchmark, we make use of OURCVFH \cite{aldoma2012our} as implemented in \cite{rusu20113d} as a shape descriptor. We use the same subset of ShapeNet \cite{chang2015shapenet} chair models used to train our SDF model and compute a shape descriptor for each chair model in the subset. During test time we compute a OURCVFH\cite{aldoma2012our} descriptor for the test model and use it to find the nearest neighbor from the subset of ShapeNet\cite{chang2015shapenet} chair models.

For registration, we first obtain a rough approximation of the optimal transform parameters. We estimate the scale as the radius of the bounding sphere for the test shape divided by the radius of the bounding sphere of its closest match from the subset of ShapeNet\cite{chang2015shapenet} chairs. We estimate the translation as the difference between the centers of their bounding spheres. Since we know the axis of rotation, we estimate the rotation by searching the space of rotation angles in increments of $20^{\circ}$. We select the rotation angle the minimizes the maximum difference between points on the test shape and their nearest neighbor on the closest matching shape. 

We refine this estimate using an algorithm inspired by the Iterative Closest Point (ICP) algorithm \cite{besl1992method}. We first sample the test shape and its closest matching shape uniformly to obtain sets of points representing each shape. Then for each iteration of the ICP algorithm, we find correspondences by solving an optimal assignment problem. We then estimate the optimal similarity transformation given the optimal correspondences. The optimal assignment problem gives us a one-to-one mapping between correspondences. This prevents us from obtaining a degenerate solution to the registration problem where the scale becomes almost zero. We make use of \cite{2020SciPy-NMeth} to solve the optimal assignment problem and \cite{umeyama1991least} as implemented in \cite{Zhou2018} to solve the registration problem.

\subsubsection{PointNet + ICP}
For this benchmark, we make use of the PointNet \cite{qi2017pointnet} classifier as implemented in \cite{kaolin2019arxiv}. We train the classifier to predict the id of the closest ShapeNet \cite{chang2015shapenet} model from our subset of chair models given an input pointcloud. During training, for each model we randomly sample 1024 point from the mesh and feed it to classifier for prediction. We add random noise to each of the points, and the set of points are re-sampled each epoch. We run it for 60 epochs after which the training loss does not significantly improve. We validate the model by making use of 30 models from the Redwood dataset \cite{choi2016large}. We do not train on the Redwood dataset \cite{choi2016large} models since they provide only a small number of models. Instead, we train on mesh models provided by our subset of ShapeNet \cite{chang2015shapenet} chair mesh models. During test time we sample 1024 points from the test shape and use it for predicting the closest matching model. 

Registration is performed using the same algorithm described in the subsection above.

\bibliographystyle{abbrv}
\bibliography{ref}

\clearpage
\afterpage{%
\begin{figure}[p]
\setlength\tabcolsep{2pt}
\centering
\begin{tabular}{ccc}
 \textbf{Test Shape} &
 \textbf{Pointnet + ICP} &
 \textbf{Ours}\\
 \includegraphics[width=0.3\textwidth]{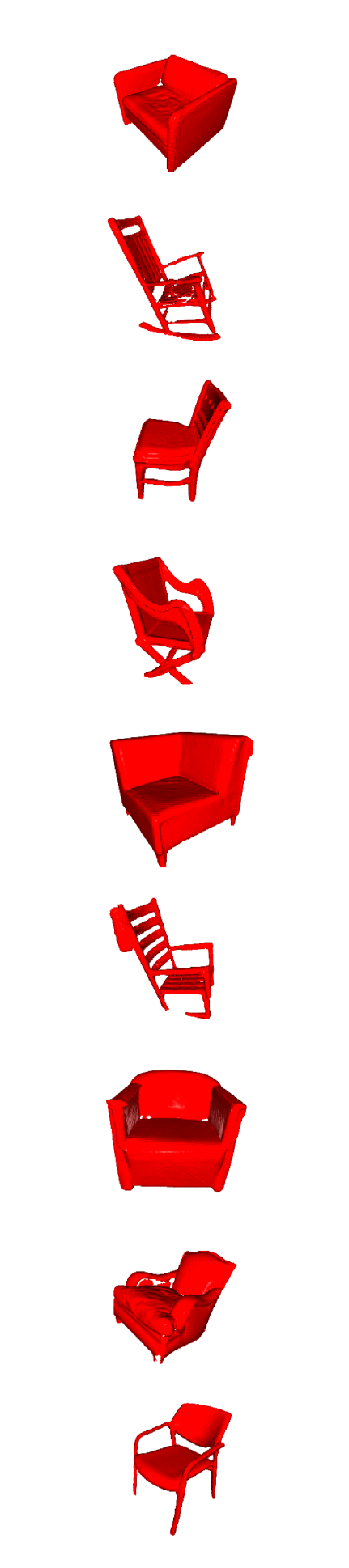}  &
 \includegraphics[width=0.3\textwidth]{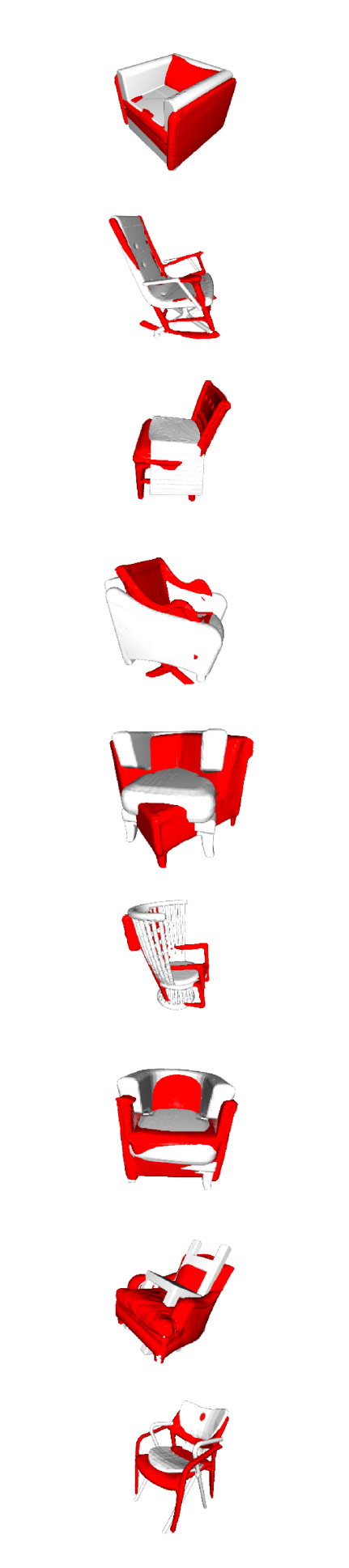} &
 \includegraphics[width=0.3\textwidth]{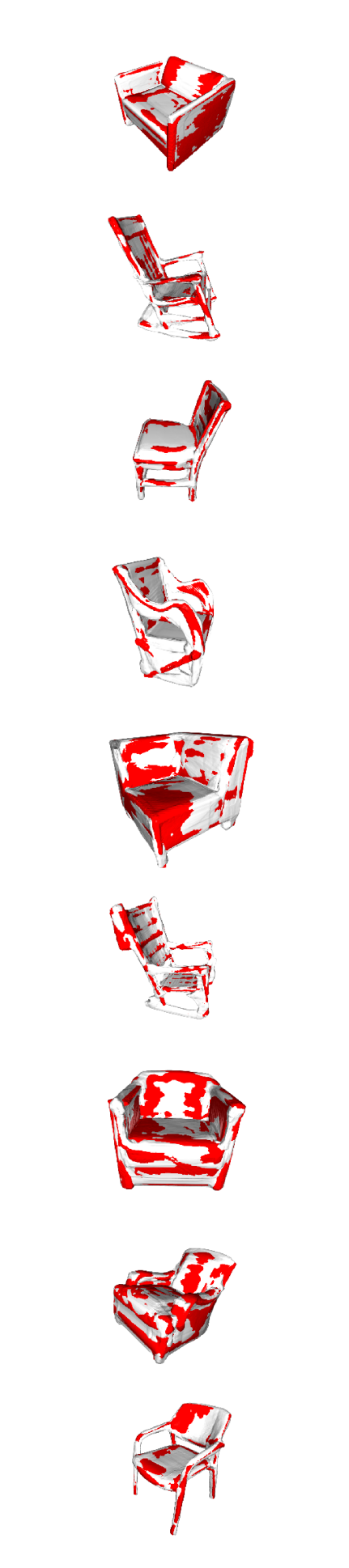} \\
\end{tabular}
\vspace{-0.2cm}
\caption{Additional qualitative result on Redwood dataset. Left to Right, test shape, results using PointNet \cite{qi2017pointnet} and ICP, our proposed method. We have shown each method's result in gray overlaid with the input mesh in red. Our algorithm provides significant qualitative improvement on par with those displayed in the main paper.}
\label{fig:supp_redwood_1}
\end{figure}
\clearpage
}
\afterpage{%
\begin{figure}[p]
\setlength\tabcolsep{2pt}
\centering
\begin{tabular}{ccc}
 \textbf{Test Shape} &
 \textbf{Pointnet + ICP} &
 \textbf{Ours}\\
 \includegraphics[width=0.3\textwidth]{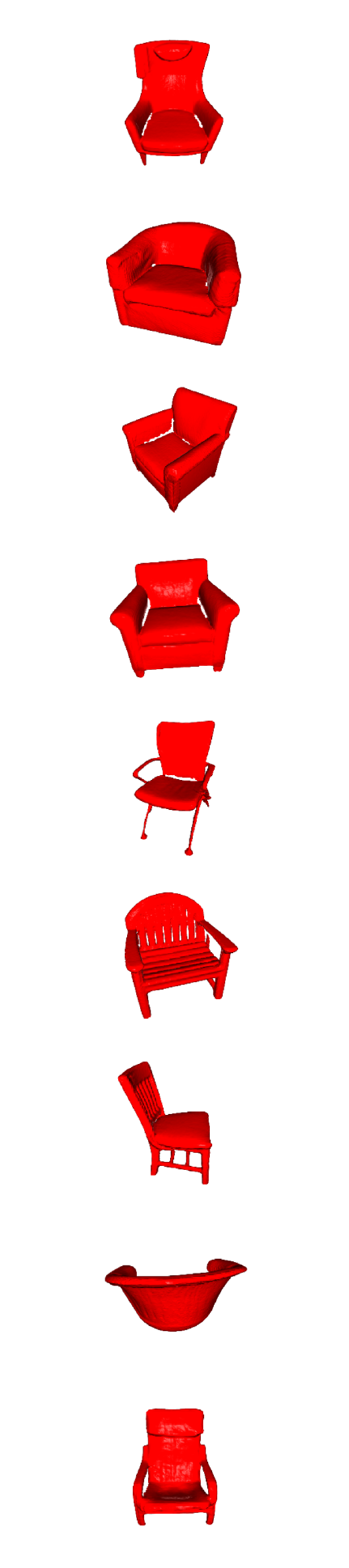}  &
 \includegraphics[width=0.3\textwidth]{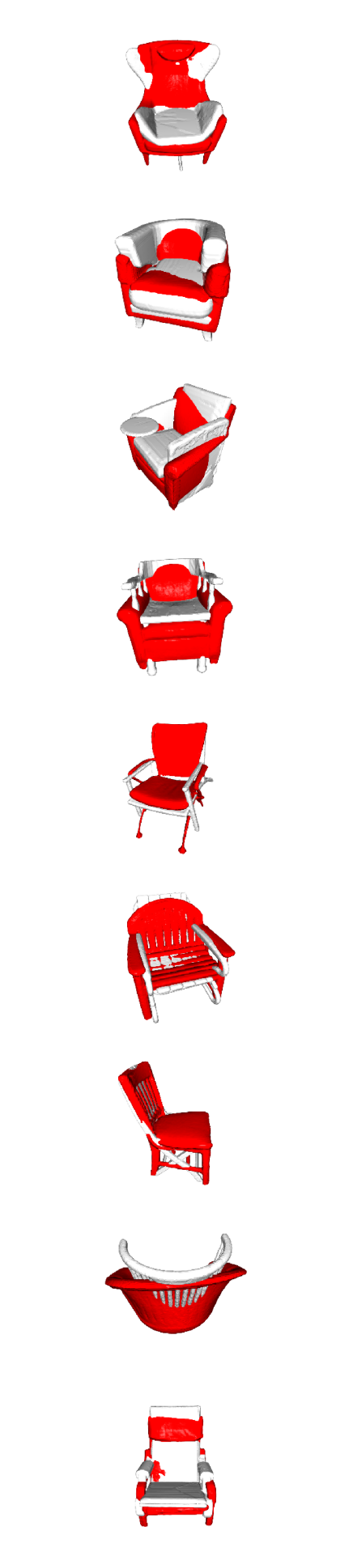} &
 \includegraphics[width=0.3\textwidth]{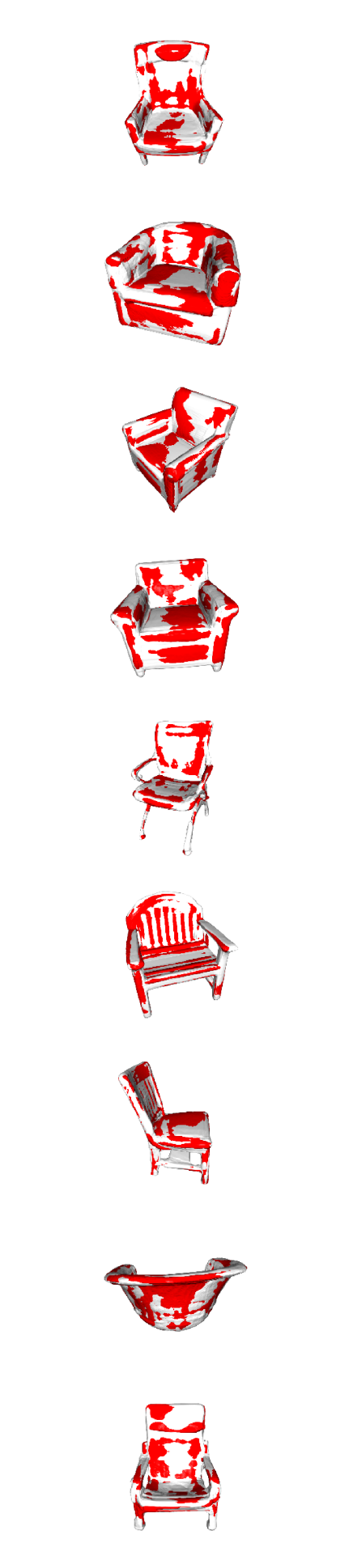} \\
\end{tabular}
\vspace{-0.2cm}
\caption{Additional qualitative result on Redwood dataset. Left to Right, test shape, results using PointNet \cite{qi2017pointnet} and ICP, our proposed method. We have shown each method's result in gray overlaid with the input mesh in red. Our algorithm provides significant qualitative improvement on par with those displayed in the main paper.}
\label{fig:supp_redwood_2}
\end{figure}
\clearpage
}

\afterpage{%
\begin{figure}[p]
\setlength\tabcolsep{2pt}
\centering
\begin{tabular}{ccc}
 \textbf{Test Shape} &
 \textbf{Pointnet + ICP} &
 \textbf{Ours}\\
 \includegraphics[width=0.3\textwidth]{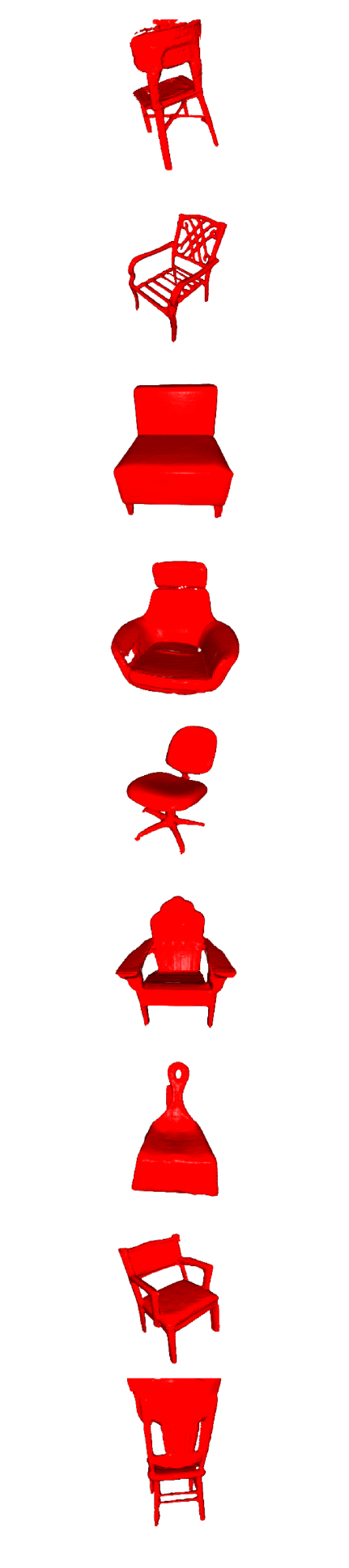}  &
 \includegraphics[width=0.3\textwidth]{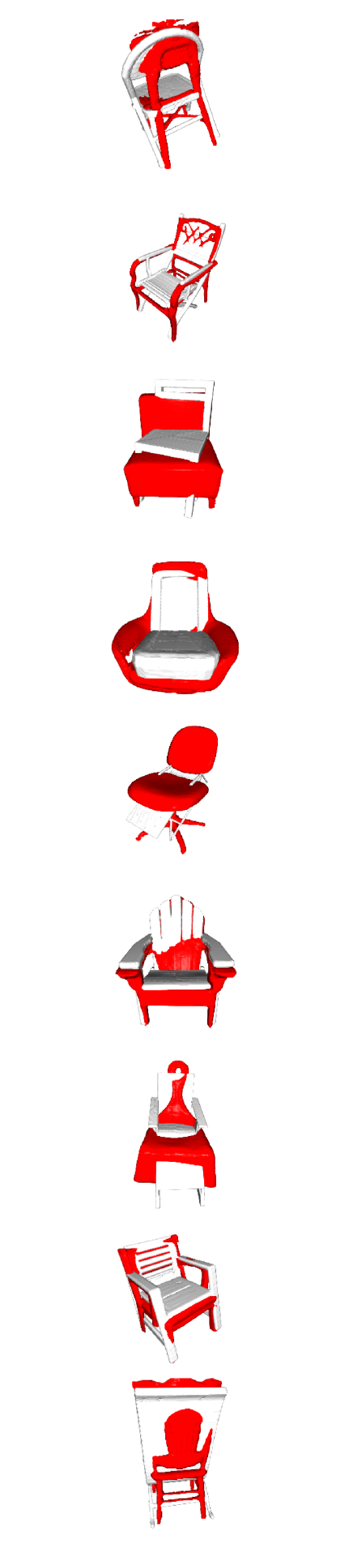} &
 \includegraphics[width=0.3\textwidth]{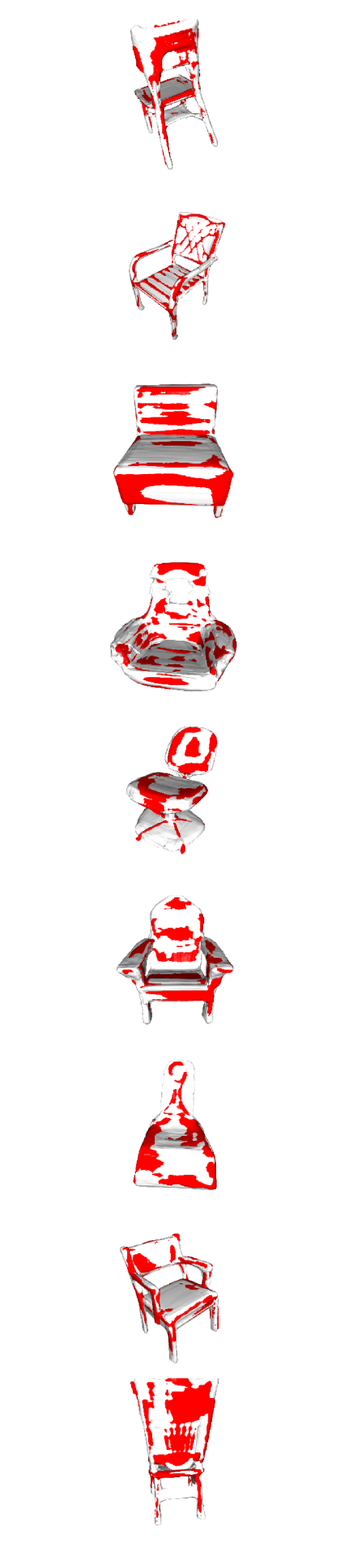} \\
\end{tabular}
\vspace{-0.2cm}
\caption{Additional qualitative result on Redwood dataset. Left to Right, test shape, results using PointNet \cite{qi2017pointnet} and ICP, our proposed method. We have shown each method's result in gray overlaid with the input mesh in red. Our algorithm provides significant qualitative improvement on par with those displayed in the main paper.}
\label{fig:supp_redwood_3}
\end{figure}
\clearpage
}

\afterpage{%
\begin{figure}[p]
\setlength\tabcolsep{2pt}
\centering
\begin{tabular}{ccc}
 \textbf{Test Shape} &
 \textbf{Pointnet + ICP} &
 \textbf{Ours}\\
 \includegraphics[width=0.3\textwidth]{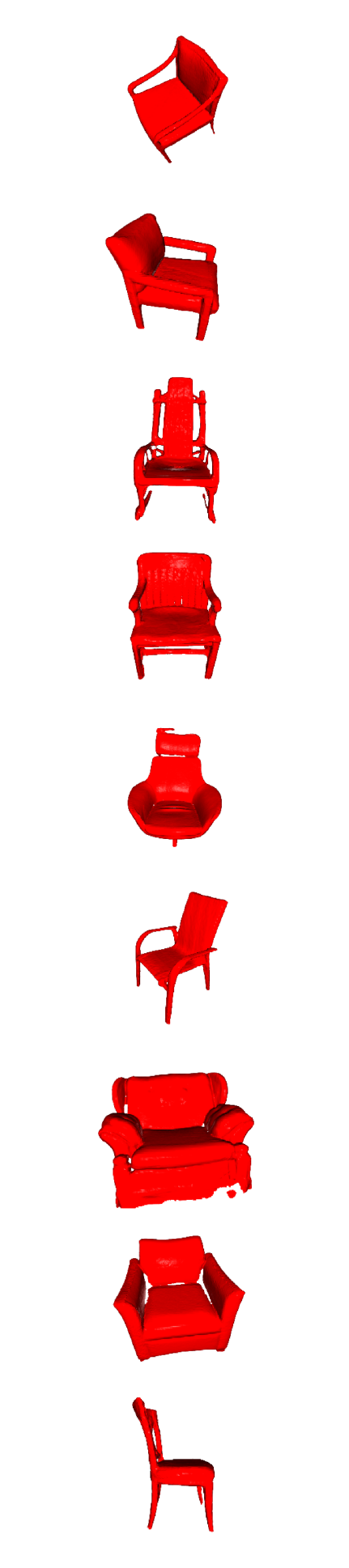}  &
 \includegraphics[width=0.3\textwidth]{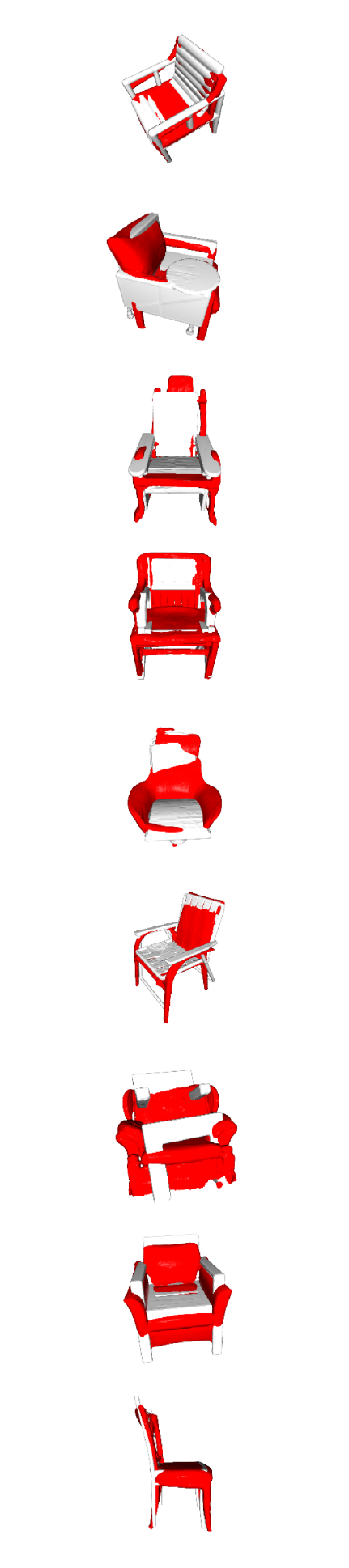} &
 \includegraphics[width=0.3\textwidth]{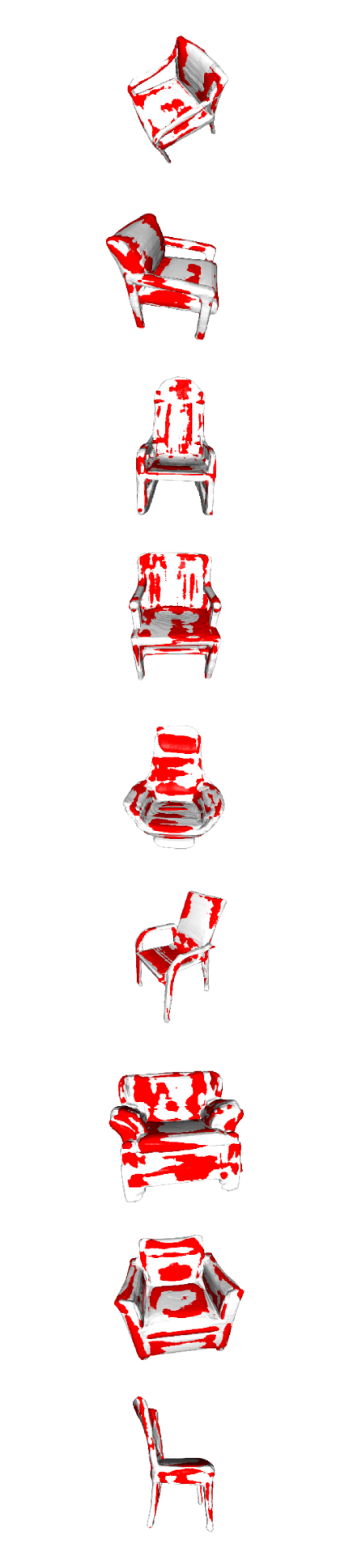} \\
\end{tabular}
\vspace{-0.2cm}
\caption{Additional qualitative result on Redwood dataset. Left to Right, test shape, results using PointNet \cite{qi2017pointnet} and ICP, our proposed method. We have shown each method's result in gray overlaid with the input mesh in red. Our algorithm provides significant qualitative improvement on par with those displayed in the main paper.}
\label{fig:supp_redwood_4}
\end{figure}
\clearpage
}

\afterpage{%
\begin{figure}[p]
\setlength\tabcolsep{2pt}
\centering
\begin{tabular}{ccc}
 \textbf{Test Shape} &
 \textbf{Pointnet + ICP} &
 \textbf{Ours}\\
 \includegraphics[width=0.3\textwidth]{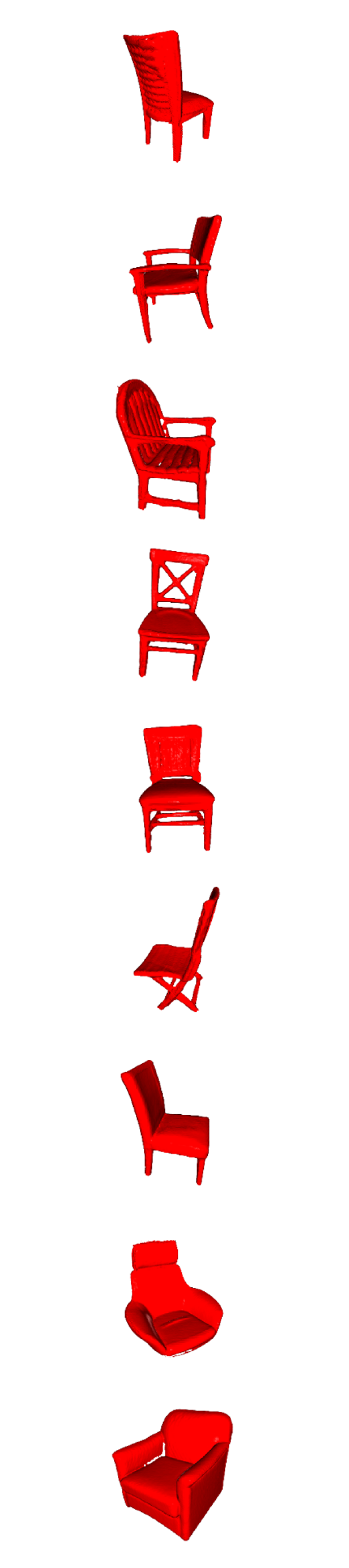}  &
 \includegraphics[width=0.3\textwidth]{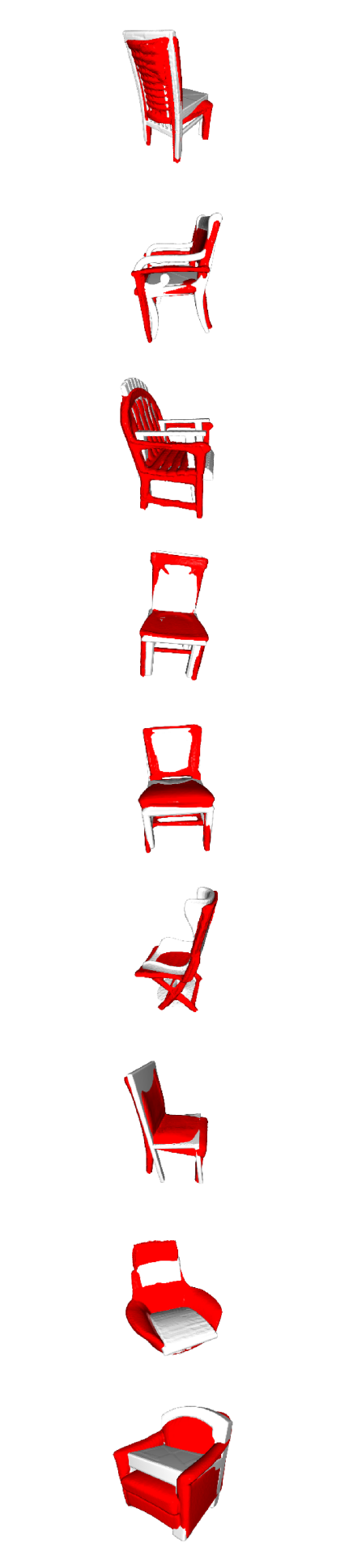} &
 \includegraphics[width=0.3\textwidth]{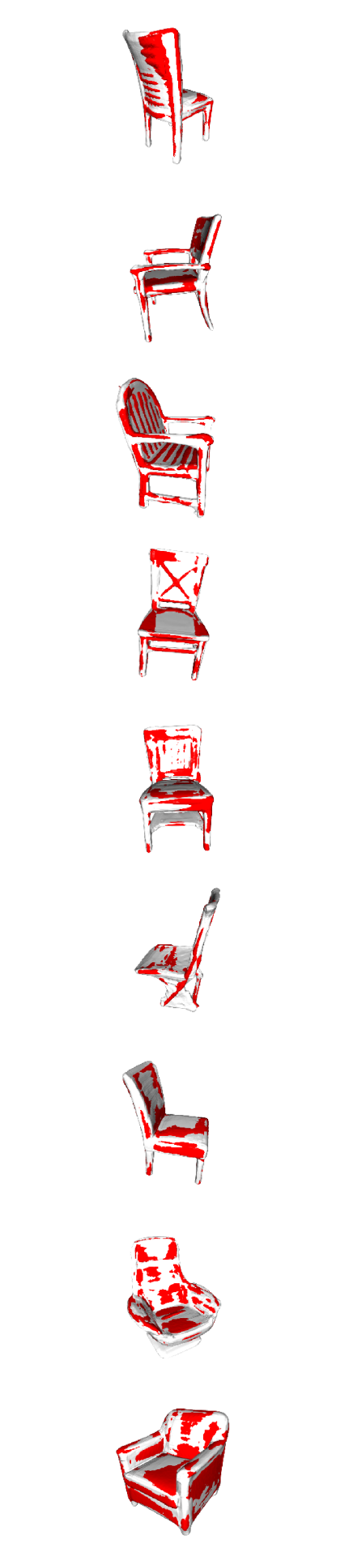} \\
\end{tabular}
\vspace{-0.2cm}
\caption{Additional qualitative result on Redwood dataset. Left to Right, test shape, results using PointNet \cite{qi2017pointnet} and ICP, our proposed method. We have shown each method's result in gray overlaid with the input mesh in red. Our algorithm provides significant qualitative improvement on par with those displayed in the main paper.}
\label{fig:supp_redwood_5}
\end{figure}
\clearpage
}

\afterpage{%
\begin{figure}[p]
\setlength\tabcolsep{2pt}
\centering
\begin{tabular}{ccc}
 \textbf{Test Shape} &
 \textbf{Pointnet + ICP} &
 \textbf{Ours}\\
 \includegraphics[width=0.3\textwidth]{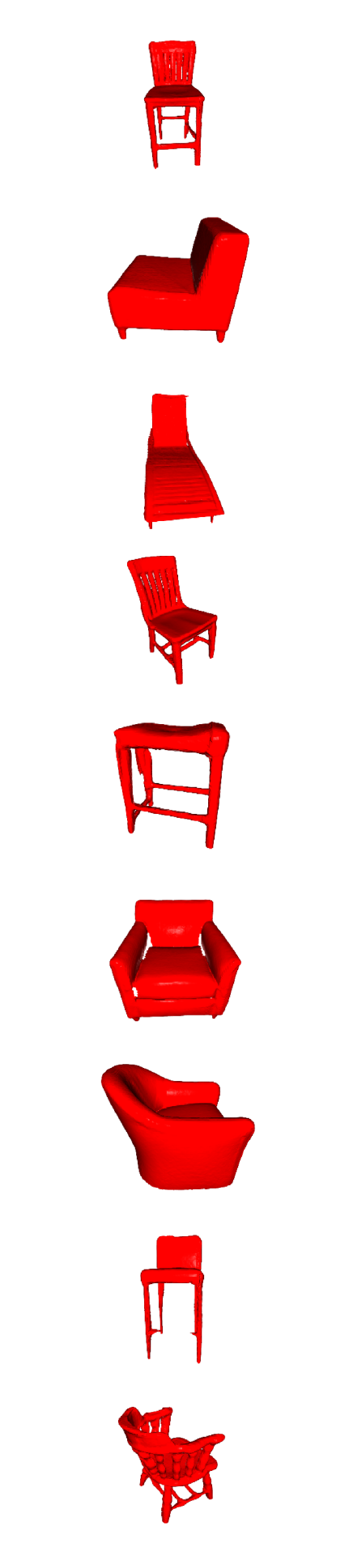}  &
 \includegraphics[width=0.3\textwidth]{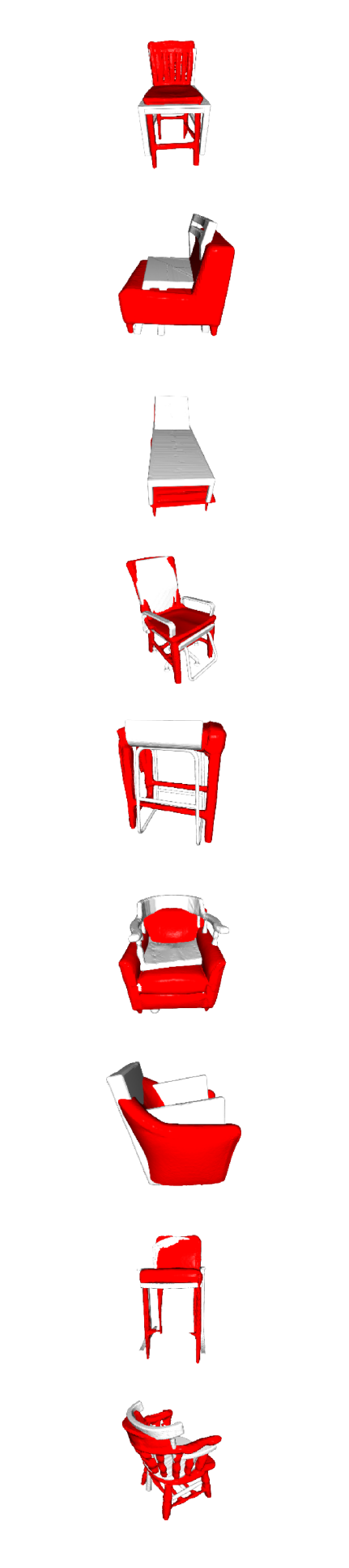} &
 \includegraphics[width=0.3\textwidth]{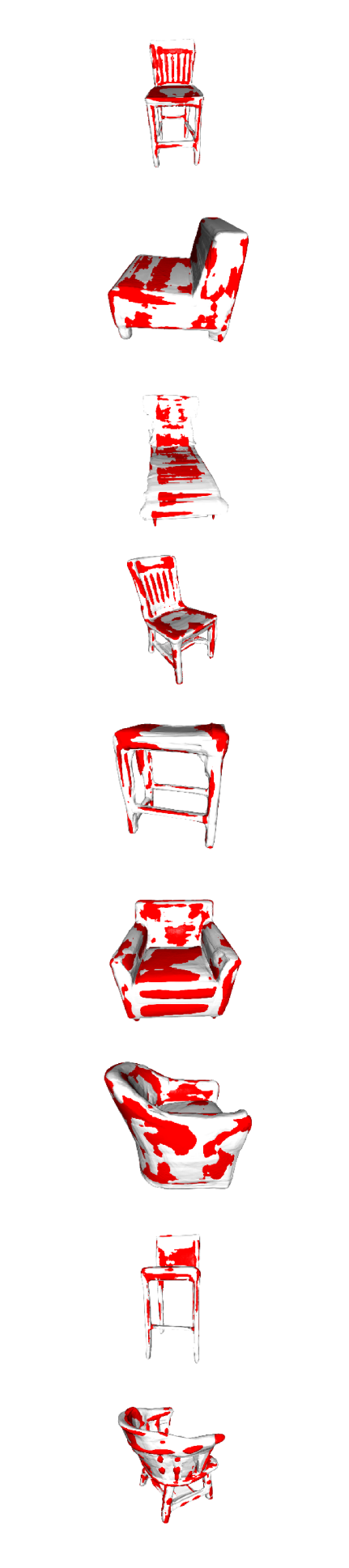} \\
\end{tabular}
\vspace{-0.2cm}
\caption{Additional qualitative result on Redwood dataset. Left to Right, test shape, results using PointNet \cite{qi2017pointnet} and ICP, our proposed method. We have shown each method's result in gray overlaid with the input mesh in red. Our algorithm provides significant qualitative improvement on par with those displayed in the main paper.}
\label{fig:supp_redwood_6}
\end{figure}
\clearpage
}

\end{document}